\documentclass[review]{elsarticle}

\journal{Neurocomputing}

\usepackage[utf8]{inputenc}
\usepackage{amsmath}        
\usepackage{amssymb}
\usepackage{tikz}
\usepackage{tabularx}
\newcolumntype{C}{>{\centering\arraybackslash}X}
\usepackage{setspace}
\usepackage{booktabs}
\newcolumntype{L}{>$l<$}
\usepackage{enumitem}
\usepackage{subfig}
\usepackage{algpseudocode}
\usepackage[colorinlistoftodos]{todonotes}
\usepackage{dsfont}
\usepackage{mathtools}
\usepackage{float}
\usepackage{bm}
\usepackage{hyperref}

\begin{document}

\begin{frontmatter}

\title{ Adversarial $\alpha$-divergence Minimization for Bayesian Approximate Inference}


\author[first_address]{Sim\'on Rodr\'iguez Santana \corref{mycorrespondingauthor}}
\cortext[mycorrespondingauthor]{Corresponding author}
\ead{simon.rodriguez@icmat.es}

\author[second_address]{Daniel Hern\'andez-Lobato}
\ead{daniel.hernandez@uam.es}

\address[first_address]{Institute of Mathematical Sciences (ICMAT-CSIC), Campus de Cantoblanco, C/Nicol\'as Cabrera, 13-15, 28049 Madrid, Spain.}
\address[second_address]{Escuela Politécnica Superior, Universidad Autónoma de Madrid, Campus de Cantoblanco, C/Franciso Tom\'as y Valiente 11, 28049 Madrid, Spain.}

\begin{abstract}

Neural networks are popular state-of-the-art models for many different tasks. They are often trained via back-propagation to find a value of the weights that correctly predicts the observed data. Although back-propagation has shown good performance in many applications, it cannot easily output an estimate of the uncertainty in the predictions made. Estimating the uncertainty in the predictions is a critical aspect with important applications, and one method to obtain this information is following a Bayesian approach to estimate a posterior distribution on the model parameters. This posterior distribution summarizes which parameter values are compatible with the data, but is usually intractable and has to be approximated. Several mechanisms have been considered for solving this problem. We propose here a general method for approximate Bayesian inference that is based on minimizing $\alpha$-divergences and that allows for flexible approximate distributions. The method is evaluated in the context of Bayesian neural networks on extensive experiments. The results show that, in regression problems, it often gives better performance in terms of the test log-likelihood and sometimes in terms of the squared error. In classification problems, however, it gives competitive results.
\end{abstract}

\end{frontmatter}

%
%

\section{Introduction}
\label{sec:Introduction}

In the past years, Neural Networks (NNs) have become very popular due to the empirical achievements in a wide variety of problems. Specifically, Deep Neural Networks (DNNs) trained with back-propagation have significantly improved the state-of-the-art in supervised learning tasks \citep{lecun2015deep}.  Moreover, variations of the simple original NN models have been specifically designed to take advantage of underlying structure on the input data. This is the case for Convolutional Neural Networks (CNNs) \cite{krizhevsky2012imagenet} or Long-Short Term Memory Networks (LSTMs) \cite{hochreiter1997long}, both of which represent some of the best performing models for dealing with structured data such as images and texts, respectively. NNs can be trained on Graphical Processing Units (GPUs), which significantly reduces the total training time and the effort needed to produce highly accurate results. These models can therefore be trained on huge amounts of data very quickly, showing excellent results in regression and a competitive performance also in classification tasks.  
In spite of the advantages described, the good performance results come with some drawbacks, such as the concerns 
about over-fitting due to the high number of parameters to be adjusted, or the lack of a confidence 
measure on the predicted outputs associated to the input data \cite{gal2016uncertainty}. More precisely, regular NNs only produce 
point-estimate predictions and do not provide any information about the certainty of such outcome. Even in multi-class 
problems where the results are given    in terms of a soft-max function which outputs probabilities, 
it is important to keep in mind that the output values do not correspond to the confidence of the 
prediction. In particular, a high class label probability may correspond to a data instance that will be 
often misclassified by the network. 

The problems described can be addressed by following a Bayesian approach in the training process, instead of relying on back-propagation for finding point-estimates of the model parameters. One of the main features of Bayesian probabilistic models such as Bayesian neural networks (BNNs) \cite{neal2012bayesian} is that they are able to capture the uncertainty in the model parameters (the network weights) and the effects it produces in 
the final predictions, therefore providing an estimate of the models' ignorance on the input data 
in each specific case. This extra output information can be used in different ways: for example, confronting problems in artificial intelligence safety, performing active learning, or dealing with possible 
adversaries which may manipulate the data \cite{gal2016uncertainty}. Summing up, uncertainty estimates 
associated to the model predictions can be very important to make optimal decisions when dealing with input 
data that the machine learning algorithm has never seen before. 

The Bayesian approach relies on computing a posterior distribution for the model parameters given the observed
data \cite{gal2016uncertainty}. This posterior distribution is obtained using Bayes' rule simply by multiplying a likelihood function (which 
captures how well specific values of the parameters explain the observed data) and a prior distribution (which includes 
prior knowledge about what potential values this parameters may take). This posterior distribution summarizes which
model parameters (\emph{i.e.}, the neural network weights) are compatible with the observed data. Intuitively, if the model 
is rather complex, the posterior will be very broad. By contrast, if the model is fairly simple, the posterior will concentrate 
on a specific region of the parameters space. The information contained in the posterior distribution can be readily translated
into a predictive distribution which carries information about the uncertainty on the predictions made. For this, one simply 
has to average the predictions of the model for each parameter configuration weighted by the corresponding posterior probability.

A difficulty of the Bayesian approach is, however, that computing the posterior distribution is intractable
for most problems. Therefore, in practice, one has to resort to approximate methods. Most of these methods approximate the exact posterior using a 
an approximate distribution $q$. 
The parameters of $q$ are tuned by minimizing a divergence between $q$ and the exact posterior. This is how methods such as variational inference (VI), expectation
propagation (EP) or black-box-alpha work in practice \citep{hernandez2015probabilistic,hernandez2016black,graves2011practical}. 
Although this methods are very fast and scalable, a limitation is the lack of flexibility of the approximate distribution $q$, 
which is often set to be a parametric distribution that cannot adequately match the exact posterior. Therefore, these methods may suffer from strong approximation bias. Importantly, a poor approximation of the exact posterior is expected to lead to a worse predictive
distribution, less accurate predictions, and a worse estimate of the uncertainty in the predictions made.

Recently, several methods have been proposed to increase the flexibility of the approximate distribution $q$
\cite{rezende2015variational,mescheder2017adversarial,liu2016stein,salimans2015markov,tran2017hierarchical}. Among these, a successful approach
is to use an implicit model for the approximate distribution $q$ \cite{li2016wild}. Under this setting, $q$ is simply obtained
by applying an adjustable non-linear function (\emph{e.g.}, given by the output of a neural network) to a source of Gaussian noise. 
If the non-linear function is flexible enough, almost any distribution can be approximated like this. The problem is, however, that even though
$q$ is a distribution that is easy to sample from, its p.d.f. can not be obtained analytically due to the complexity of the
non-linear function. This makes approximate inference (\emph{i.e.}, tuning the parameters of the non-linear function) very challenging. 
Adversarial variational Bayes (AVB) is a technique that solves this problem \cite{mescheder2017adversarial}. AVB minimizes the 
Kullback-Leibler (KL) divergence between $q$ and the exact posterior. This technique avoids evaluating the p.d.f. of $q$ 
by learning a discriminator network that estimates the log-ratio between the posterior approximation $q$ and the prior 
distribution over the model parameters. 

AVB and also other methods such as VI or EP (only locally and in the reversed way) rely on minimizing the KL divergence between 
the approximate distribution $q$ and the exact posterior. The $\alpha$-divergence generalizes the KL divergence and includes a parameter 
$\alpha \in (0,1]$ that can be adjusted. In particular, when $\alpha \rightarrow 0$, the $\alpha$-divergence tends 
to the KL-divergence optimized by VI. By contrast, if $\alpha=1$, the $\alpha$-divergence is the reversed KL-divergence, \emph{i.e.}, 
the KL-divergence between the exact posterior and $q$, which is locally optimized by EP. Recently, it has been empirically shown that one 
can obtain better results in terms of the approximate predictive distribution by minimizing $\alpha$-divergences locally using intermediate 
values of the $\alpha$ parameter, in the case of parametric $q$ \cite{hernandez2016black}. It is not clear however if one can also obtain
better results in the case of implicit models for $q$, such as the one considered by AVB. 

In this paper we extend AVB to locally minimize $\alpha$-divergences, in an approximate way, 
instead of the regular KL divergence, being $\alpha$ a parameter. 
Therefore, this method can be seen as a generalization of AVB that allows to
optimize a more general class of divergences, resulting in flexible approximate 
distributions $q$ with different properties. When $\alpha \rightarrow 0$, the proposed 
method converges to standard AVB. When $\alpha = 1$ the proposed method 
is similar to EP with a flexible approximate distribution $q$.  We have evaluated such a 
method in the context of Bayesian Neural Networks and tested different values 
of the $\alpha$ parameter. The experiments carried out involve several regression and classification problems extracted from the 
UCI repository. They show that in regression problems one can obtain, in general, better prediction results than those of AVB and 
standard VI, in terms of the mean squared error and the test log-likelihood, by using intermediate values of $\alpha$.
In classification problems the proposed approach is competitive with AVB.

\section{Variational Inference and Adversarial Variational Bayes}
 \label{sec:AVB}

Adversarial Variational Bayes (AVB) is an extension of variational inference (VI) \citep{beal2003variational} that
allows for implicit models for the approximate distribution $q$. We describe here first VI and then AVB in detail.  

\subsection{Variational Inference}

Let $\mathbf{w}$ be the latent variables of the model, \emph{e.g.},
the neural network weights. The task of interest in VI is to approximate the posterior distribution of $\mathbf{w}$ given the observed 
data. For simplicity we will focus on regression models, but the method is broadly applicable to any model and is not limited to neural 
networks. 

Consider a training set $\mathcal{D}=\{\mathbf{x}_i,y_i\}_{i=1}^N$, where $\mathbf{x}_i$ is some $d$-dimensional input 
vector and $y_i \in \mathds{R}$ is the associated label. The posterior distribution is given by Bayes' rule:
\begin{align}
p(\mathbf{w}|\mathcal{D}) &= \frac{p(\mathbf{y}|\mathbf{w},\mathbf{X}) p(\mathbf{w})}{p(\mathcal{D})} = 
	\frac{\left[ \prod_{i=1}^N p(y_i|\mathbf{w},\mathbf{x}_i) \right]p(\mathbf{w})}{p(\mathcal{D})}\,,
	\label{eq:posterior}
\end{align}
where $\mathbf{X}$ is a matrix with the observed vectors of input attributes and $\mathbf{y}=(y_1,\ldots,y_N)^\text{T}$.
Furthermore we have assumed i.i.d. data and hence, the likelihood  factorizes as 
$p(\mathbf{y}|\mathbf{w},\mathbf{X})=\prod_{i=1}^N p(y_i|\mathbf{w},\mathbf{x}_i)$.
In (\ref{eq:posterior}) $p(\mathbf{w})$ is the prior distribution of the latent variables of the model (\emph{i.e.}, the neural network weights) and 
$p(\mathcal{D})=\int p(\mathbf{y}|\mathbf{w},\mathbf{X}) p(\mathbf{w}) d\mathbf{w}$ is just a normalization constant.
In the case of regression problems $p(y_i|\mathbf{w},\mathbf{x}_i)$ is often a Gaussian distribution, \emph{i.e.}, $\mathcal{N}(y_i|f(\mathbf{x}_i), \sigma^2)$,
where $f(\mathbf{x}_i)$ is the output of the neural network and $\sigma^2$ is the variance of the output noise. Furthermore, $p(\mathbf{w})$ is often a factorizing
Gaussian with zero mean and variance $\sigma_0^2$ (see  \emph{e.g.}, \cite{hernandez2015probabilistic,graves2011practical,hernandez2016black}).
Given (\ref{eq:posterior}) the predictive distribution of the model for the label $y^\star$ of a new test point $\mathbf{x}_\star$ is:
\begin{align}
	p(y^\star|\mathcal{D}) &= \int p(y^\star|\mathbf{w},\mathbf{x}^\star) p(\mathbf{w}|\mathcal{D}) d \mathbf{w}\,.	
	\label{eq:predictive}
\end{align}
The model prediction would be the expected value of $y^\star$ under (\ref{eq:predictive}) and the confidence in the prediction can be estimated, \emph{e.g.}, 
by the standard deviation.  In practice, $p(\mathbf{w}|\mathcal{D})$ is intractable because $p(\mathcal{D})$ has no closed form expression and one has to use an approximation to this 
distribution in (\ref{eq:predictive}). 

VI approximates (\ref{eq:posterior}) using a parametric distribution $q(\mathbf{w})$ which is often a factorizing Gaussian 
$\mathcal{N}(\mathbf{w}|\bm{\mu},\bm{\Sigma})$ with $\bm{\Sigma}$ a diagonal matrix. Let $\phi$ be the set of parameters of $q(\mathbf{w})$, \emph{i.e.}, $\phi=\{\bm{\mu},\bm{\Sigma}\}$. 
These parameters are adjusted to minimize the KL divergence between $q(\mathbf{w})$ and the exact posterior (\ref{eq:posterior}). 
Consider the following decomposition of $\log p(\mathcal{D})$:
\begin{align}
\log p(\mathcal{D}) &= \mathds{E}_{q_\phi(\mathbf{w})}[\log p(\mathbf{y},\mathbf{w}|\mathbf{X}) - \log q(\mathbf{w})] + \text{KL}(q(\mathbf{w})||p(\mathbf{w}|\mathcal{D})) 
	\,,
	\label{eq:decomposition}
    \end{align}
where $\text{KL}(q(\mathbf{w})||p(\mathbf{w}|\mathcal{D}))$ is the KL divergence between $q(\mathbf{w})$ and the exact posterior:
\begin{align}
	\text{KL}(q(\mathbf{w}) || p(\mathbf{w}|\mathcal{D})) &= - \int q_\phi(\mathbf{w}) \log \frac{p(\mathbf{w}|\mathcal{D})}{q_\phi(\mathbf{w})} d \mathbf{w} \geq 0\,.
\end{align}
The KL divergence is always non-negative and is only zero if the two distributions are the same. Therefore, by minimizing this divergence VI enforces that $q(\mathbf{w})$
looks similar  to the exact posterior (\ref{eq:posterior}). 

Because $\log p(\mathcal{D})$ is a constant term independent of $\phi$, the KL divergence between $q(\mathbf{w})$ and the exact posterior can be simply 
minimized by maximizing the first term in the r.h.s. of (\ref{eq:decomposition}) with respect to $\phi$. This term is often referred to as
the evidence lower bound:
\begin{equation}
\begin{aligned}
\mathcal{L}(\phi) &= \mathds{E}_{q_\phi(\mathbf{w})}[\log p(\mathbf{y},\mathbf{w}|\mathbf{X}) - \log q(\mathbf{w})] 
	 \\ & = \sum_{i=1}^N \mathds{E}_{q_\phi(\mathbf{w})}[p(y_i|\mathbf{w},\mathbf{x}_i)] - \text{KL}(q(\mathbf{w})||p(\mathbf{w})) \,,
	\label{eq:lower_bound_vi}
\end{aligned}
\end{equation}
where $\text{KL}(q(\mathbf{w})||p(\mathbf{w}))$ is the KL divergence between $q(\mathbf{w})$ and the prior $p(\mathbf{w})$. 
If $q(\mathbf{w})$ and the prior are Gaussian, there is a closed form expression for this 
divergence. The maximization of (\ref{eq:lower_bound_vi})
can be done using stochastic optimization techniques that sub-sample the training 
data and that approximate the required expectations using Monte Carlo samples (see \cite{graves2011practical} for further details). The hyper-parameters of 
the model, \emph{i.e.}, the noise and prior variance $\sigma^2$ and $\sigma_0^2$
are estimated by maximizing $\mathcal{L}(\phi)$, which approximates 
$\log p(\mathcal{D})$ since $\text{KL}(q(\mathbf{w})||p(\mathbf{w}|\mathcal{D}))$ is expected to be fairly small.
Finally, after training, the posterior approximation can replace the exact posterior in (\ref{eq:predictive})
and the predictive distribution for new data can be approximated by a 
Monte Carlo average over the posterior samples.

\subsection{Adversarial Variational Bayes}
\label{sec:avb}

AVB extends VI to account for implicit models for the approximate distribution $q(\mathbf{w})$. An implicit model for $q(\mathbf{w})$ is a distribution that 
is easy to generate samples from, but that lacks a closed form expression for the p.d.f. An example is a source of standard Gaussian 
noise that is non-linearly transformed by a neural network. That is,
\begin{align}
	q_\phi(\mathbf{w}) = \int \delta \left(\mathbf{w} -\mathbf{f}_\phi(\bm{\epsilon})\right) \mathcal{N}(\bm{\epsilon}|\mathbf{0},\mathbf{I}) d \bm{\epsilon}\,,
	\label{eq:implicit}
\end{align}
where $\mathbf{f}_\phi(\bm{\epsilon})$ is the output of a neural network that receives $\bm{\epsilon}$ at the input and $\delta(\cdot)$ is a delta function.
In general, the integral in (\ref{eq:implicit}) is intractable due to the strong non-linearities of the neural network. Nevertheless, it is very easy to
generate $\mathbf{w} \sim q_\phi$. For this, one only has to generate $\bm{\epsilon}\sim \mathcal{N}(\mathbf{0},\mathbf{I})$ to then compute $\mathbf{w}=\mathbf{f}_\phi(\bm{\epsilon})$.
If the noise dimension is large enough and $\mathbf{f}_\phi(\cdot)$ is flexible enough, any probability distribution can be described like this.
Therefore, implicit models can alleviate the approximation bias of VI with parametric distributions $q(\mathbf{w})$.

Using an implicit distribution in VI is challenging because the lower bound in (\ref{eq:lower_bound_vi}) cannot be easily 
evaluated nor optimized. The reason is that the term $\text{KL}(q(\mathbf{w})||p(\mathbf{w}))$, \emph{i.e.}, the KL divergence between the approximate distribution $q(\mathbf{w})$ and the prior requires the p.d.f. of $q(\mathbf{w})$. AVB provides an elegant 
solution to this problem. The aforementioned term can be written as:
\begin{align}
	\text{KL}(q(\mathbf{w})||p(\mathbf{w})) = \mathds{E}_{q_\phi(\mathbf{w})}\left[ \log q_\phi(\mathbf{w}) - \log p(\mathbf{w}) \right] = 
	\mathds{E}_{q_\phi(\mathbf{w})}\left[ T(\mathbf{w}) \right]\,,
	\label{eq:KL_q_prior}
\end{align}
where $T(\mathbf{w})$ is simply the log-ratio between $q_\phi$ and the prior. AVB proposes to estimate this log-ratio as the output of another neural network
that discriminates between samples of $\mathbf{w}$ generated from $q_\phi$ and from the prior \cite{mescheder2017adversarial}.
This technique has also been considered in other works \citep{tran2017hierarchical,huszar2017variational,li2016wild}.
Let $T_\omega(\cdot)$ be the output of the discriminator. The following objective is considered for optimizing the discriminator assuming $q_\phi(\mathbf{w})$
is fixed:
\begin{align}
\label{eq:objective_discriminator_general}
    \max_{\omega} \quad \mathds{E}_{q_\phi(\mathbf{w})} \left[\log \sigma (T_\omega(\mathbf{w})) + \mathds{E}_{p(\mathbf{w})} [\log (1 - \sigma (T_\omega(\mathbf{w})))] \right]\,,
\end{align}
where $\sigma(\cdot)$ is the sigmoid-function. Roughly speaking, this objective tries to make the discriminator differentiate 
between samples generated from $q_\phi(\mathbf{w})$ and from the prior $p(\mathbf{w})$.

If the discriminator $T_\omega$ is considered flexible enough to represent any function of $\mathbf{w}$, it is possible to prove that the optimal 
discriminator behaves as expected by providing the correct log-ratio between the two distributions. If we rewrite (\ref{eq:objective_discriminator_general}) making explicit the dependence on both $q_\phi(\mathbf{w})$ and $p(\mathbf{w})$  we obtain
\begin{align}
\label{eq:discriminator_objective_integral}
    \max_{\omega} \int \left[ q_\phi(\mathbf{w}) \log \sigma (T_\omega(\mathbf{w})) + p(\mathbf{w}) \log (1 - \sigma (T_\omega(\mathbf{w})) \right] d\mathbf{w}\,.
\end{align}
This integral is maximal for $T_\omega(\mathbf{w})$ if and only if the integrand is maximal for every $\mathbf{w}$ value. The shape of the integrand is:
\begin{align}
    a \log t + b \log (1-t),
\end{align}
for $a = q_\phi(\mathbf{w})$, $b = p(\mathbf{w})$, and $t = T_\omega(\mathbf{w})$.
Its maximum value is attained at $t = \frac{a}{a+b}$. Therefore, the optimal solution $T_{\omega^\star}$ is
\begin{align}
\sigma(T_{\omega^\star}(\mathbf{w})) & = \frac{q_\phi (\mathbf{w})}{q_\phi(\mathbf{w}) + p(\textbf{w})},    
\end{align}
or equivalently,
\begin{align}
    T_{\omega^\star} (\mathbf{w}) & = \log q_\phi(\mathbf{w}) - \log p(\mathbf{w}),
\end{align}
which is the result desired to correctly estimate the KL divergence between $q_\phi$ and the prior.
In particular, the discriminator can be plugged in (\ref{eq:KL_q_prior}) and the expectation can be 
approximated simply by a Monte Carlo average by generating samples from $q_\phi$.

Given $T_{\omega^\star}$ the lower bound employed in AVB is obtained by re-writing the evaluation of the KL 
divergence between $q_\phi$ and the prior:
\begin{align}
\mathcal{L}(\phi) & = \sum_{i=1}^N \mathds{E}_{q_\phi(\mathbf{w})}[p(y_i|\mathbf{w},\mathbf{x}_i)]  - \mathds{E}_{q_\phi(\mathbf{w})}[T_{\omega^\star}(\mathbf{w})] \,.
	\label{eq:lower_bound_avb}
\end{align}
Note that all the required expectations can be simply approximated by generating samples from $q_\phi$ and the sum across the training data
can be approximated using mini-batches. This lower bound can be hence easily maximized w.r.t. $\phi$ using stochastic optimization techniques.
For this, however, we need to differentiate the stochastic estimate with respect to $\phi$. This may seem complicated since 
$T_{\omega^\star}(\mathbf{w})$ is defined as the solution of an auxiliary optimization problem that depends on $\phi$. 
However, due to the expression for the optimal discriminator, it can be showed that 
$\mathds{E}_{q_\phi(\mathbf{w})} \left( \nabla_{\phi} T_{\omega^\star} (\mathbf{w}) \right) = 0$. Therefore 
the dependence of $T_{\omega^\star}(\mathbf{w})$ w.r.t $\phi$ can be ignored. See \cite{mescheder2017adversarial} for further details. 
In practice, both $q_\phi$ and the discriminator $T_\omega(\mathbf{w})$ are trained simultaneously. However, $q_\phi$ is updated by maximizing (\ref{eq:lower_bound_avb}) using a smaller learning rate than the one used to update the discriminator $T_{\omega}$, which considers the objective in (\ref{eq:objective_discriminator_general}). This helps achieving that $T_{\omega}$ is an accurate estimator of the log-ratio between $q_\phi$ and the prior, and that 
the KL divergence is correctly estimated when updating $q_\phi$.

\subsection{Adaptive Contrast}
\label{sec:adaptive_contrast}

AVB relies on a good approximation $T_\omega(\mathbf{w})$ to the optimal discriminator. 
Although in the non-parametric limit this is achieved, in practice $T_\omega(\mathbf{w})$ 
can fail to be sufficiently close to the optimal discriminator. This a consequence of 
AVB calculating the discriminator between  $q_\phi$, the posterior approximation and the prior, 
which are often very different distributions. This results in practice in a more \textit{relaxed} 
performance of the estimated discriminator, which has no problem telling apart samples from one density 
or the other, but that fails to correctly estimate the log-ratio between probability distributions. 

In \cite{mescheder2017adversarial} a solution is proposed, which consists in introducing a new 
auxiliary conditional probability distribution $r_\alpha (\mathbf{w})$ with known density that 
approximates $q_{\phi}$. This auxiliary distribution is set to be a factorizing Gaussian whose mean and variances match those of $q_\phi$. Using this extra distribution, 
the objective in (\ref{eq:lower_bound_vi}) is rewritten as 
\begin{align}
\label{eq:auxiliar_mod_objective_AVB}
\mathcal{L}(\phi) &=  - \text{KL}(q_{\phi}(\mathbf{w})||r_\alpha(\mathbf{w})) + \mathds{E}_{q_{\phi}(\mathbf{w})} \left[ 
	\log p(\mathbf{y}| \mathbf{w}, \mathbf{X}) + \log p(\mathbf{w}) - \log r_{\alpha} (\mathbf{w}) \right]\,.
\end{align}
If $r_{\alpha} (\mathbf{w})$ approximates well $q_{\phi} (\mathbf{w})$, the KL divergence between 
these distributions will often be much smaller than $\text{KL}(q_{\phi}(\mathbf{w})  || p(\mathbf{w}) )$, which 
facilitates learning the correct probability ratio. 

This technique is called \textit{adaptive contrast}, because the divergence is not being calculated 
between $q_\phi$ and the prior, but between $q_\phi$ and the adaptive distribution $r_\alpha$. 
Therefore, the discriminator now estimates $\text{KL}(q_{\phi}(\mathbf{w})||r_\alpha(\mathbf{w}))$ and 
hence the log-ratio between $q_\phi$ and $r_\alpha$. More precisely, introducing this new auxiliary distribution, the lower bound becomes
\begin{align}
\mathcal{L}(\phi) &= 
      \mathds{E}_{q_{\phi} (\mathbf{w})} \left[ - T_\omega(\mathbf{w}) - \log r_{\alpha} (\mathbf{w}) + 
	\log p(\mathbf{y}| \mathbf{w}, \mathbf{X}) + \log p(\mathbf{w}) \right]\,,
\end{align}
where now $T_\omega(\mathbf{w})$ approximates the optimal discriminator between 
samples from $r_{\alpha}(\mathbf{w})$ and $q_{\phi} (\mathbf{w})$. 
Moreover, the KL divergence in (\ref{eq:auxiliar_mod_objective_AVB}) 
is invariant under any change of variables. Therefore, it can be rewritten as:
\begin{align}
     \text{KL}(q_{\phi}(\mathbf{w})||r_\alpha(\mathbf{w})) & = \text{KL} (\tilde{q}_{\phi}(\mathbf{w})|| r_0(\mathbf{w}))\,,
\end{align}
where $\tilde{q}_{\phi}(\tilde{\mathbf{w}})$ is the distribution of 
the standardized vector $\tilde{\mathbf{w}}$, whose $j$-th component is
given by $\tilde{w}_j \coloneqq \frac{w_j - \mu_j}{\sqrt{\Sigma_{j,j}}}$ 
(with $\mu_j$ and $\Sigma_{j,j}$ the mean and variance of $w_j$, respectively), 
and $r_0 (\tilde{\mathbf{w}})$ is a standard Gaussian distribution. 
Therefore, the discriminator $T_\omega(\mathbf{w})$ just needs to look for differences between samples from the 
normalized posterior approximation and from a standard Gaussian distribution.  
The mean and variances of $\mathbf{w}$ under $q_{\phi}$ can simply be estimated using samples from this distribution.

\section{Alpha Divergence Minimization}
 \label{sec:Alpha_divergence_BNNs}

Before describing the proposed method, we briefly review here the $\alpha$-divergence, of which 
we make extensive use. Let $p$ and $q$ be two distributions over the vector $\bm{\theta}$.
The $\alpha$-divergence between $p$ and $q$ is non-negative and only equal to 
zero if $p=q$ \cite{amari2012differential}. The corresponding expression is given by
\begin{align}
    D_\alpha [p|q] & = \frac{1}{\alpha (1 - \alpha )} 
	\left( 1 - \int p(\bm{\theta})^\alpha q(\bm{\theta})^{1 - \alpha} d\bm{\theta} \right)\,.
\end{align}
This divergence has a parameter $\alpha \in \mathds{R} \setminus \{0,1\}$.
Depending on the value of $\alpha$ it recovers different well-known divergences between probability distributions. 
For example, 
\begin{align}
    D_1 [p|q] & = \lim_{\alpha \rightarrow 1 } D_\alpha [p|q] = \text{KL} [p||q]\,, \label{eq:KL_pq} \\
    D_0 [p|q] & = \lim_{\alpha \rightarrow 0 } D_\alpha [p|q] = \text{KL} [q||p]\,, \label{eq:KL_qp} \\
    D_{\frac{1}{2}} [p|q] & = 2 \int \left( \sqrt{p(\boldsymbol{\theta})} -  
	\sqrt{q(\boldsymbol{\theta})}\right)^2 d\boldsymbol{\theta} = 4\text{Hel}^2[p|q]\,.  \label{eq:Hellinger_distance}
\end{align}
The first two limiting cases given by (\ref{eq:KL_pq}) and (\ref{eq:KL_qp}) represent the two different 
possibilities for the KL-divergence between distributions. Moreover, (\ref{eq:Hellinger_distance}) is 
known as the \textit{Hellinger distance}, which is the only instance in the family of $\alpha$-divergences 
which is symmetric between both distributions. 

\begin{center}
\begin{figure}[h]
    \includegraphics[width=0.99\textwidth]{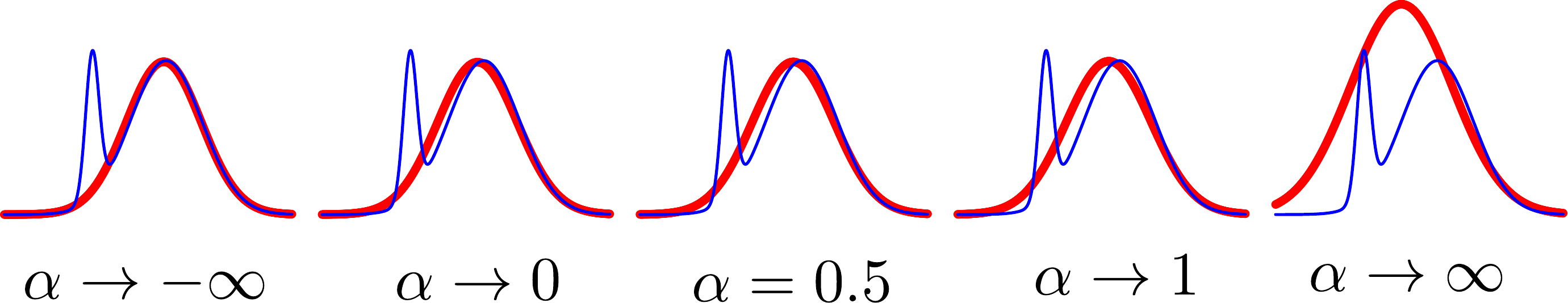}
    \caption{Changes on the approximate distribution $q$ (in red) when trying to 
	approximate it to the original distribution $p$ (in blue) using different values for $\alpha$ in the $\alpha$-divergence. 
	When $\alpha \rightarrow -\infty$ the approximate distribution tries to cover a local mode of the target distribution (\textit{exclusive distribution}). When
	$\alpha \rightarrow \infty$ the approximate distribution tries to cover the whole target distribution (\textit{inclusive distribution}).
	}
    \label{fig:example_gaussians_alpha_div}
\end{figure}
\end{center}

The value of the $\alpha$ parameter in the $\alpha$-divergence has a strong impact in the
inference results. Thus, to further understand its effect 
let us consider a toy problem in which we try to approximate a \emph{slightly complex} 
distribution $p$ with a simpler one, $q$. If we considered for example 
$p$ as a bimodal distribution and $q$ as a simple Gaussian distribution we 
would obtain the results displayed in Figure \ref{fig:example_gaussians_alpha_div} 
(reproduced from \cite{minka2005divergence}). In this figure, the resulting 
unnormalized approximating distributions exhibit different behaviors (the expression for the $\alpha$-divergence can be generalized so that it can be evaluated on distributions that need not be normalized, see \cite{minka2005divergence} for further details). First of all, in the limit of $\alpha \rightarrow -\infty$, $q$, here represented in red, tends to cover only the mode with the larger mass of the two present in $p$. By contrast, when $\alpha \rightarrow \infty$, $q$ tends to cover the whole $p$ distribution, overlaying the latter completely. This can be seen in terms of the form of the $\alpha$-divergence. More precisely, for $\alpha \leq 0$, the $\alpha$-divergence emphasizes $q$ to be small whenever $p$ is small (thus it could be considered as zero-forcing). On the other hand, when $\alpha \geq 1$, it can be said that the divergence is inclusive, following the terminology of \cite{frey2001sequentially}. In this case, the divergence enforces $q > 0$ wherever $p > 0$, hence avoiding not having probability density in regions of the input space in which $p$ takes large values. 

In rest of the cases, $\alpha$ lays inside the interval $(0,1)$. 
The behavior of $q$ is intermediate between the two extreme 
possibilities that we have seen so far. In 
Figure \ref{fig:example_gaussians_alpha_div} we can see that when 
$\alpha \rightarrow 0$ the $q$ distribution is more centered in the main mode of $p$, 
whereas in $\alpha \rightarrow 1$ it begins to open to account for some of the mass 
of the secondary peak of $p$. This behavior also happens when the distributions 
being considered are more complex than these ones, and therefore one has to 
be careful when choosing $\alpha$. In particular, the optimal value of $\alpha$ may 
depend on the task at hand and the particular model one is working with. As it has been 
pointed out before, when $\alpha$ is restricted to be in the interval $(0,1)$ we can obtain two notable
results at the extremes, $D_\alpha = \text{KL} (q||p)$ for $\alpha \rightarrow 0$ and $D_\alpha = \text{KL} (q||p)$ 
for $\alpha \rightarrow 1$. These two expressions are directly related to two of 
the main methods for approximate inference, Variational Inference \cite{beal2003variational} and 
Expectation Propagation \cite{minka2001expectation}, respectively.  
 
\section{Adversarial Alpha Divergence Minimization}
\label{sec:Adversarial_alpha_divergence_minimization}

So far we have seen that AVB is a flexible method for approximate inference that 
allows for the use of implicit models for the approximate distribution $q_\phi$.
If the implicit model is complex enough, AVB should be able to capture the features 
of the target distribution. However, AVB strongly relies on the KL divergence to enforce that the approximate distribution looks similar to the target distribution. 
In Section \ref{sec:Alpha_divergence_BNNs} we have pointed out that by employing a more general 
form of divergence one can obtain more flexible results, which depending on the task may mean a better balance between approximating a local mode of the posterior distribution (\textit{exclusive distribution}) or having high probability density in
all the regions of the input space in which the target distribution has high probability (\textit{inclusive distribution}). 
The method proposed here is a generalization of AVB that allows for optimizing 
in an approximate way the $\alpha$-divergence, instead of the KL divergence. By changing the $\alpha$ parameter one 
can hence obtain different approximate distributions $q(\mathbf{w})$ to the ones resulting before. We refer to this method as Adversarial
Alpha Divergence Minimization. Our assumption here is that, if we are able to use values of $\alpha$ different from the ones that 
are used in AVB (\emph{i.e.}, $\alpha \rightarrow 0$), we can perhaps obtain different approximating distributions 
that yield better results in terms of the prediction error or the test log-likelihood, since it will allow the system to balance the importance assigned to mode-selecting and all-covering behaviors.  

As discussed earlier, when $\alpha \rightarrow 0 $, the $\alpha$-divergence recovers the KL divergence typical from VI and AVB, 
and when $\alpha \rightarrow 1$, the opposite KL divergence is restored (which is the one employed in other algorithms such 
as Expectation Propagation \cite{minka2001expectation}). The mid range of values of alphas between $0$ and $1$ remains to be explored here. Therefore, we are going to search for intermediate $\alpha$ values more suited for each learning task, and besides this, we can also try 
to analyze the general behavior of the approximate inference method depending on the selection of this parameter.

To introduce the use of $\alpha$-divergences in the context of AVB we modify the AVB objective function so 
that it accounts for this extra parameter as well. To do so we follow the approach described in 
\cite{li2017dropout}, which allows for the approximate minimization of $\alpha$-divergences with
approximate distributions $q(\mathbf{w})$ that are not implicit. To make this description more complete, we will first describe briefly the 
power expectation propagation objective function \cite{minka2004}. We will also make a brief introduction to black-box $\alpha$, an extension of power EP on which we have based our approach on.  
    
\subsection{Power Expectation Propagation}
    
We consider the objective function of a general method for approximate inference known as 
power expectation propagation (PEP) \citep{minka2004}. PEP allows for minimizing $\alpha$-divergences in an
approximate way, but constrains the approximate distribution $q(\mathbf{w})$ to belong to the family of exponential 
distributions (\emph{e.g.}, $q(\mathbf{w})$ is a Gaussian distribution). More precisely, the global minimization 
of the $\alpha$-divergence is intractable, except when $\alpha \rightarrow 0$ (see \citep{minka2005divergence} 
for further details). Let the unnormalized target distribution be the product of several factors,
\emph{i.e.}, $p\propto \prod_i f_i$. If we have i.i.d. data, this is always the case, since
the likelihood factorizes. PEP approximates $p(\mathbf{w}|\mathcal{D})$ by $q(\mathbf{w})$, which is written as a product of simple
factors $q \propto \prod_i \tilde{f}_i$. Each $\tilde{f}_i$ belongs to the exponential 
family (\emph{e.g.} a Gaussian factor) and approximates the corresponding exact factor $f_i$. PEP minimizes the $\alpha$-divergence locally, instead of globally.  
In particular, PEP minimizes the $\alpha$-divergence between the tilted distributions of the model, and the 
approximate distribution $q(\mathbf{w})$. The tilted distributions of the model are those distributions in which one 
approximate factor $\tilde{f}_i$ is replaced by the corresponding exact factor $f_i$. Namely, 
$p^{\setminus j} \propto f_j \prod_{i\neq j} \tilde{f}_i$. Therefore, PEP minimizes 
$D_\alpha[p^{\setminus j}|q]$ for all $j$. In general, it is expected that a 
local minimization of the $\alpha$-divergence gives similar results to a global minimization while being a much simpler problem,
as indicated in \citep{minka2005divergence}. 

To perform the local minimization of the $\alpha$-divergence PEP optimizes the following objective function:
\begin{align}
\mathcal{L}(\phi, \{\theta_i\}_{i=1}^N) &= 
	\log Z_{q} - \log Z_{p(\mathbf{w})}
	+ \frac{1}{\alpha} \sum_{i=1}^N \log \mathds{E}_{q_\phi(\mathbf{w})} 
	\left[ \left( \frac{p(y_i|\mathbf{w},\mathbf{x}_i)}{\tilde{f}_i(\mathbf{w})} 
	\right)^\alpha \right]\,,
	\label{eq:pep_objective}
\end{align}
where $Z_{q}$ is the normalization constant of $q_\phi$, $Z_{p(\mathbf{w})}$ is the normalization constant of the prior,
$\phi$ are the parameters of $q(\mathbf{w})$ and $\{\theta_i\}_{i=1}^N$ are the parameters of 
the approximate factors $\tilde{f}_i$. 
In this case, we have assumed that the prior distribution need not be
approximated and already belongs to the exponential family (\emph{i.e.}, it is a Gaussian prior). 
When $\alpha \rightarrow 0$, (\ref{eq:pep_objective}) converges to the lower bound of 
VI in (\ref{eq:lower_bound_vi}) \cite{minka2005divergence}. Therefore, a local minimization of 
the KL divergence employed in VI is equivalent to a global minimization.

In practice, PEP solves the problem $\max_\phi\,\, \min_{\{\theta_i\}_{i=1}^N} \mathcal{L}(\phi, \{\theta_i\}_{i=1}^N)$,
which is a complicated task since it requires a slow double loop algorithm \cite{heskes2002expectation}.
Furthermore, PEP does not scale to big data since it maintains an approximate factor associated to each 
likelihood factor, which results in a space complexity of $\mathcal{O}(N)$.

\subsection{Black-box $\alpha$-divergence Minimization}

Black-box-$\alpha$ (BB-$\alpha$) is an improvement over the previous method, PEP, that addresses some
of its limitations like the memory space requirements and also allows to make approximate inference 
on complicated probabilistic models \cite{hernandez2016black}. For this, the PEP objective function is
rewritten as:
\begin{align}
\mathcal{L}(\phi) &= 
	\log Z_q - \log Z_{p(\mathbf{w})}
	+ \frac{1}{\alpha} \sum_{i=1}^N \log \mathds{E}_{q_\phi(\mathbf{w})} 
	\left[ \left( \frac{p(y_i|\mathbf{w},\mathbf{x}_i)}{\tilde{f}(\mathbf{w})} 
	\right)^\alpha \right]\,,
	\label{eq:bba_objective}
\end{align}
where now there is only one approximate factor $\tilde{f}(\mathbf{w})$ that is replicated $N$ times, one per
each complicated likelihood factor. Therefore, $q_\phi(\mathbf{w}) \propto \tilde{f}(\mathbf{w})^N p(\mathbf{w})$. 
This solves the problem of having to store in memory the parameters of $N$ factors. Furthermore, there is a 
one to one map between $\tilde{f}(\mathbf{w})$ and $q_\phi(\mathbf{w})$. This means that the max-min optimization problem 
of PEP is transformed into just a standard maximization problem (w.r.t to the parameters of $q(\mathbf{w})$, $\phi$), 
which can be solved using standard optimization techniques. Importantly, the expectations 
in (\ref{eq:bba_objective}) can be approximated via Monte Carlo sampling
and the sum across the training data can be approximated using a mini-batch. The consequence is that BB-$\alpha$
scales to big datasets, as (\ref{eq:bba_objective}) can be optimized using stochastic techniques, and moreover, 
it can be applied to complicated probabilistic models (\emph{e.g.}, Bayesian neural networks) in which the 
required expectations are intractable. Again, when $\alpha \rightarrow 0$ (\ref{eq:bba_objective}) converges 
to the lower bound of VI in (\ref{eq:lower_bound_vi}). When $\alpha = 1$, (\ref{eq:bba_objective}) is approximately
equal to the objective optimized by Expectation Propagation \citep{hernandez2016black}.
A limitation of BB-$\alpha$ is, however, that the approximate distribution $q(\mathbf{w})$ is restricted to be 
inside the exponential family. This is because it must be written as the product of an approximate factor
times the prior distribution. That is, $q_\phi(\mathbf{w}) \propto \tilde{f}(\mathbf{w})^N p(\mathbf{w})$.
This is a major limitation that makes difficult using implicit models for $q(\mathbf{w})$.

\subsection{Reparameterization of the Black-box-$\alpha$ Objective}
\label{sec:reparametrization_bba}

In this section we will do an analogous reparametrization for the general expression of the BB-$\alpha$ objective suggested in \cite{li2017dropout} for VI, but we will instead apply it to AVB. This reparametrization will allow to approximately minimize $\alpha$-divergences with flexible distributions $q(\mathbf{w})$ such as the ones resulting from implicit models. Therefore, it will enable us to complete the formulation of our model by making use of this type of divergences, extending on the existing formulation of AVB. To this end, first consider the following alternative expression for the BB-$\alpha$ objective:
\begin{align}
    \mathcal{L}_\alpha (\phi) & = \frac{1}{\alpha} \sum_{i=1}^N \log 
	\mathds{E}_{q_\phi(\mathbf{w})} \left[ \left(\frac{p(y_i| \mathbf{x}_i, \mathbf{w}) 
	p(\mathbf{w})^{1/N}}{q_\phi(\mathbf{w})^{1/N}}\right)^\alpha \right]\,.
\label{eq:general_BBalpha_energy_function}
\end{align}
In this expression we observe that the hypothesis that $q_\phi(\mathbf{w}) \propto \tilde{f}^N(\mathbf{w}) p(\mathbf{w})$
is not required anymore (both $Z_q$ and $\tilde{f}(\mathbf{w})$ are removed from the expression) and $q(\mathbf{w})$ can 
be an arbitrary distribution. It is possible to show that (\ref{eq:general_BBalpha_energy_function})
and (\ref{eq:bba_objective}) become equivalent if $q(\mathbf{w})$ belongs to the exponential family \cite{li2017dropout}.
A difficulty is, however, that this expression requires the evaluation of the density $q_\phi(\mathbf{w})$ which in practice can be hard to compute.

To overcome the limitation described before, in \cite{li2017dropout}, they reparametrize 
(\ref{eq:general_BBalpha_energy_function}) using the so-called \emph{cavity distribution}. 
That is, the distribution given by the ratio $q_\phi / \tilde{f}^\alpha$.  
If $\tilde{q}_\phi(\mathbf{w})$ denotes a free-form cavity distribution, the posterior 
approximation $q_\phi$ is given by: 
\begin{align}
    q_\phi(\mathbf{w}) & = \frac{1}{Z_q} \tilde{q}_\phi(\mathbf{w}) \left( \frac{\tilde{q}_\phi(\mathbf{w})}{p(\mathbf{w})}  \right)^{\frac{\alpha}{N-\alpha}} 
\end{align}
where we assume $Z_q < +\infty$ is the normalizing constant to make $q(\mathbf{w})$ 
a valid distribution. When $\alpha / N \rightarrow 0$ we have that 
$q \rightarrow \tilde{q}$ (and $Z_q \rightarrow 1$ by assumption), 
and this is the case either if we choose $\alpha \rightarrow 0$ or for a sufficiently large $N$ (i.e. $N \rightarrow +\infty$), see \cite{li2017dropout}. We rewrite now (\ref{eq:general_BBalpha_energy_function}) in terms of $\tilde{q}$ rather than $q(\mathbf{w})$:
\begin{align}
	\mathcal{L}_\alpha(\phi) &= 
	\frac{1}{\alpha} \sum_{i=1}^N \log \int 
	\left( \frac{1}{Z_q} \tilde{q}_{\phi}(\mathbf{w}) \left( 
	\frac{\tilde{q}_{\phi}(\mathbf{w})}{p(\mathbf{w})} \right)^\frac{\alpha}{\alpha - N} \right)^{1 - \frac{\alpha}{N}}
	p(\mathbf{w})^\frac{\alpha}{N} p(y_i|\mathbf{w},\mathbf{x}_i)^\alpha d \mathbf{w} \nonumber \\
	&= -\frac{N}{\alpha} \left( 1 - \frac{\alpha}{N} \right) 
	\log \int \tilde{q}_\phi(\mathbf{w}) \left( 
	\frac{\tilde{q}_\phi(\mathbf{w})}{p(\mathbf{w})} \right)^\frac{\alpha}{N- \alpha} 
	d \mathbf{w} \nonumber \\ 
	& \quad + \frac{1}{\alpha} \sum_{i=1}^N \log \mathds{E}_{\tilde{q}_{\phi}(\mathbf{w})}
	\left[p(y_i | \mathbf{x}_i, \mathbf{w})^\alpha \right] \nonumber \\
        &= \frac{1}{\alpha} 
	\sum_{i=1}^N \log \mathds{E}_{\tilde{q}_\phi(\mathbf{w})} \left[
	p(y_i | \mathbf{x}_i, \mathbf{w})^\alpha \right]
	-\text{R}_\beta [\tilde{q} | p] 
	\,, 
\end{align}
where $\beta = N / (N - \alpha)$ and $\text{R}_\beta [\tilde{q} | p] $ represents 
the \emph{R\'enyi divergence} of order $\beta$ \cite{renyi1961measures}, which is defined as
\begin{align}
    \text{R}_\beta [q | p] & = \frac{1}{\beta - 1} \log \int \tilde{q}(\mathbf{w})^\beta p(\mathbf{w})^{1 - \beta} d \mathbf{w}.
\end{align}

Importantly, when $\alpha / N \rightarrow 0$ we recover $q \rightarrow \tilde{q}$ and
$\mathcal{L}_\alpha(\phi)$ converges to the objective of VI. 
Also, we have that $\text{R}_\beta [\tilde{q} | p] \rightarrow \text{KL} [\tilde{q} || p] = \text{KL} [q || p]$  
if $\text{R}_\beta [\tilde{q} | p] < +\infty$ (which is true assuming $Z_q < +\infty$ and $\alpha / N \rightarrow 0 $).
Therefore, when this quotient tends to zero, we can make further approximations 
for the BB-$\alpha$ energy function as described in (\ref{eq:general_BBalpha_energy_function}), finally obtaining
\begin{align}
    \label{eq:alpha_avb_energyfunction}
    \mathcal{L}_\alpha (\phi) \approx
	\frac{1}{\alpha} 
	\sum_{i=1}^N \log \mathds{E}_{q_\phi(\mathbf{w})}[p(y_i | \mathbf{x}_i, \mathbf{w})^\alpha]
	- \text{KL}[q_{\phi} (\mathbf{w})||p(\mathbf{w})]\,     .  
\end{align}
This will be the objective function that we will optimize in our approach. Note that the expectations 
in (\ref{eq:alpha_avb_energyfunction}) can be estimated via Monte Carlo sampling. In particular, 
$\log \mathds{E}_{q_\phi(\mathbf{w})}[p(y_i | \mathbf{x}_i, \mathbf{w})^\alpha] \approx \log [ K^{-1} \sum_{k=1}^K 
p(y_i | \mathbf{x}_i, \mathbf{w}_k)^\alpha]$, for $K$ samples of $\mathbf{w}$ drawn from $q_\phi$.
Of course, this estimate is biased, as a consequence of the non-linearity of the $\log(\cdot)$ function, 
however, the bias can be controlled with $K$. Furthermore, we expect a 
similar behavior as in standard BB-$\alpha$, in which the bias has been shown to be very small even for 
$K=10$ samples. See \cite{hernandez2016black} for further details.

The objective in (\ref{eq:alpha_avb_energyfunction}) has been obtained under some conditions that need not be
true in practice, \emph{e.g.} the quotient $\alpha / N \rightarrow 0$ (\emph{i.e.}, either $\alpha$ is small, $N$ is sufficiently large or a combination of both). Nevertheless, it is much simpler to estimate and optimize than the one in
(\ref{eq:general_BBalpha_energy_function}). It is also similar to the objective functions found in the deep learning 
bibliography (\emph{i.e.}, a loss function plus some regularizer, \emph{i.e.}, the KL divergence), 
but it still maintains the qualities of an approximate Bayesian inference algorithm. Importantly, 
(\ref{eq:alpha_avb_energyfunction}) allows for implicit models for $q_\phi$. The only term that is difficult to approximate is 
$\text{KL}[q_{\phi} (\mathbf{w})||p(\mathbf{w})]$. However, the approach described in Section \ref{sec:avb} can be used
for that purpose. By changing the $\alpha$ parameter of the method we will be able to interpolate
between AVB ($\alpha \rightarrow 0$) and an EP like algorithm ($\alpha = 1$). 
Note that when $\alpha \rightarrow 0$, (\ref{eq:alpha_avb_energyfunction}) is expected to focus on reducing the training error since 
the factor $\alpha^{-1} \log \mathds{E}_{q_\phi(\mathbf{w})}[p(y_i | \mathbf{x}_i, \mathbf{w})^\alpha]$ will converge
to $\mathds{E}_{q_\phi(\mathbf{w})}[\log p(y_i | \mathbf{x}_i, \mathbf{w})]$, with $p(y_i | \mathbf{x}_i, \mathbf{w})$ 
typically a Gaussian distribution
with mean given by the output of the neural network and noise variance $\sigma^2$. By contrast,
when $\alpha = 1$, (\ref{eq:alpha_avb_energyfunction}) will be expected to focus more on the training log-likelihood.
Intermediate values of $\alpha$ will trade-off between these two tasks, which may lead to better 
generalization properties.

With respect to the specific details of the architecture of the proposed approach, it consists of a structure analogous to the one presented in AVB \cite{mescheder2017adversarial}. Therefore, its structure can be divided into three networks: An implicit model for $q_\phi$, which takes as input Gaussian noise and outputs neural network weight samples $\mathbf{w}$ from the approximated weights posterior distribution (\emph{i.e.}, the generator network); a discriminator, which estimates the KL term present in (\ref{eq:alpha_avb_energyfunction}) as done in \cite{mescheder2017adversarial}; and finally the main network, that uses the samples of the weights generated previously to evaluate the factor $p(y_i|\mathbf{x}_i,\mathbf{w})$. The whole system is optimized all together. Furthermore, any potential hyper-parameter (\emph{e.g.}, the  prior variance $\sigma_0^2$ or the output noise variance $\sigma^2$) is tuned simply by maximizing (\ref{eq:alpha_avb_energyfunction}).

Finally, as a last remark concerning the implementation of the proposed method, we have also included as trainable parameters both the mean and variances of the Gaussian noise which is used as input in the generator network (the implicit model for the weights, $q_\phi (\mathbf{w})$, in (\ref{eq:implicit})), $\bm{\epsilon} \sim \mathcal{N}(\bm{\mu}_\text{noise}, \bm{\Sigma}_\text{noise})$, with $\bm{\Sigma}_\text{noise}$ a diagonal matrix. This allows for a more expressive implicit model for $q_\phi (\mathbf{w})$, since it increases its flexibility by allowing the tuning of the broad parameters that control its input. Using this, the model is expected to reproduce to a higher degree of accuracy the original posterior distribution of the model parameters (neural network weights). 

\subsubsection{Annealing Factor}

The proposed method, as described, may suffer from convergence to bad local optima. More precisely, it can pay
too much attention from the beginning to the KL term, failing to explain the observed data. Therefore, it is 
convenient to bias the training of the method in such a way that, at least at the beginning, it does not consider 
relevant the KL term. If that is the case, it will not try to make the approximate distribution look like the prior
during the first steps of the optimization process, hopefully avoiding bad local optima.

In order to accomplish this, we incorporated the technique described in \cite{sonderby2016ladder}. Using this as an example, we define a sufficiently large \emph{warm-up} period for which we will train our 
model \emph{turning on progressively} the KL term in the objective function. We do this 
by changing slightly the original formulation of (\ref{eq:alpha_avb_energyfunction}) to 
introduce an extra \textit{annealing parameter} $\beta$. That is,
\begin{align}
\label{eq:ladder_alpha_avb_energyfunction}
    \mathcal{L}_\alpha (\phi) & \simeq \frac{1}{\alpha} \sum_{i=1}^N \log \mathds{E}_{q_\phi(\mathbf{w})}
	[p(y_i | \mathbf{x}_i, \mathbf{w})^\alpha] - \beta \text{KL}[q_{\boldsymbol{\phi}} (\mathbf{w})||p(\mathbf{w})]\,,
\end{align}
where $\beta$ starts being equal to $0$ and grows linearly to 
$1$ during a certain number of epochs. In every experiment where this modification has been implemented, the number of warm-up epochs is 
selected to be the 10\% of the total epochs assigned for training the algorithm in a given dataset (mostly those extracted from the UCI repository). 
We have observed that this significantly improves the results of the proposed method.
In the case of the synthetic problems, the warm up period is set to 
500 epochs from a total number of 3000 epochs. In the experiments with big data
we have not included the annealing factor since it has not been observed to be beneficial.

\section{Related Work}
 \label{sec:Related_work}
 
Obtaining the uncertainty in the predictions of machine learning
algorithms is a widely spread problem. Originally, this problem has been addressed
either by sampling-based methods or by optimization-based  methods
\cite{li2016wild}. In sampling-based methods, the posterior distribution is approximated by drawing 
samples from the exact posterior to then use these for inference and prediction. For this, a Markov chain 
is run, whose stationary distribution coincides with the target distribution. On the other hand, 
optimization-based methods introduce an approximate distribution $q(\mathbf{w})$ whose 
parameters are adjusted to match the exact posterior through the optimization of a certain objective. 

Each of the approaches described has advantages and disadvantages. Sampling methods can be unbiased only 
asymptotically, and moreover they can be highly computationally expensive since the Markov chain 
has to be run for long time in practice. Similarly, optimization-based techniques are usually limited by 
the definition of the approximating distribution, which is often parametric, and therefore they 
may lack expressiveness. Two examples of these methods are Markov chain Monte Carlo (MCMC) in 
the case of sampling-based methods \cite{duane1987hybrid,neal2011mcmc,neal2012bayesian}, and variational inference (VI)
or expectation propagation (EP) in the case of optimization-based methods \cite{minka2001expectation,
jordan1999introduction,graves2011practical,beal2003variational, soudry2014expectation}. The method proposed here alleviates 
some of the problems of these two techniques. Specifically, it allows for flexible approximate 
distributions and it also scales to large datasets, whereas in some of these cases, large datasets can be a burden to deal with \cite{hoffman2017learning}.

Most modern techniques for approximate inference take advantage 
of the speed of optimization-based methods and try to preserve the flexibility of
sampling-based methods with the goal of obtaining the best results possible in terms of 
computational cost and accuracy of the approximation. There are, however, many different ways of combining 
both approaches, which is showcased by the wide variety of methods 
proposed. In this section we review some of them. Nevertheless, almost all of them rely on
optimizing the KL divergence between $q(\mathbf{w})$ and the target distribution. The approach proposed
by us is more general and can minimize a collection of divergences known as the $\alpha$-divergence, 
which includes also the KL divergence as a particular case.

One example is the work in \cite{titsias2018unbiased}, where 
it is described how to estimate the gradient of the VI objective
when using an implicit model for the approximate distribution $q(\mathbf{w})$.
For this, the method described in that work relies in a combination 
of Markov chain Monte Carlo methods and VI. While this approach seems promising,
its implementation is very complicated since it relies on running an inner Markov 
chain inside of the optimization process of the approximate distribution $q(\mathbf{w})$. This
Markov chain has also parameters that need to be correctly adjusted and that may depend 
on the probabilistic model.

Another approach that allows for flexible approximate distributions $q(\mathbf{w})$ within the context of VI 
is normalizing flows (NF) \cite{rezende2015variational}. In NF one starts with a simple parametric
approximate distribution $q(\mathbf{w})$ whose samples are modified using parametric non-linear invertible 
transformations.  If these transformations are chosen carefully, the p.d.f. of the resulting distribution 
can be evaluated in closed form, avoiding the problems arising from the use of implicit models for $q(\mathbf{w})$. 
The problem of NF is that the family of transformations that can be used is limited, which may 
constrain the flexibility of the approximate distribution $q(\mathbf{w})$. 

\emph{Stein Variational Gradient Descent}, proposed in \cite{liu2016stein}, is a general VI method 
that consists in transforming a set of \emph{particles} to match the exact posterior distribution. 
The results obtained are shown to be competitive with other state-of-the-art methods, but the main 
drawback here is that there is a computational bottleneck on the number of particles that need to 
be stored to accurately represent the posterior distribution. More precisely, this method lacks a 
way to generate samples from the approximate distribution $q(\mathbf{w})$. The number of samples is fixed 
initially, and these are optimized by the method.

The work in \cite{salimans2015markov} combines VI and MCMC methods to obtain flexible approximate
posterior distributions. The key concept is to use a Markov chain as the approximate distribution $q(\mathbf{w})$
in VI. The parameters of this chain can then be adjusted to match as close as possible the target
distribution in terms of the KL divergence. This is an interesting idea. However, it is also limited
by the difficulty of evaluating the p.d.f. of the approximate distribution. This is solved in \cite{salimans2015markov}
by learning a backward model, that infers the p.d.f. of the initial state of the Markov chain given the generated
samples. Learning this backward model accurately is a complex task and several simplifications are introduced that
may affect the results.

Another approach used for approximate inference in the context of neural networks is Probabilistic Back-propagation \cite{hernandez2015probabilistic}. This method 
computes a forward propagation of probabilities through the neural network to then do 
back-propagation of the gradients. Although it has been proven to be a fast approach with high 
performance, it is limited by the expressiveness of the posterior approximation.
In particular, the approximate distribution is restricted to be Gaussian.
This means that this method will suffer from strong approximation bias.
The same applies to a standard application of VI in the context of Bayesian neural 
networks \cite{graves2011practical}.

The minimization of $\alpha$-divergences in the context of Bayesian neural networks has also been
addressed in \cite{hernandez2015probabilistic}. In that work it is described Black-box-$\alpha$, a method for 
approximate inference that allows for very complex probabilistic models and that is efficient and allows
for big datasets. The main limitation is, however, that the approximate distribution $q(\mathbf{w})$ must
belong to the exponential family. That is, the approximate distribution has to be Gaussian,
and hence, this method will also suffer from approximation bias. Therefore, Black-box-$\alpha$ is expected to be 
sub-optimal when compared to the method proposed in this paper, which allows for implicit models in the
approximate distribution $q(\mathbf{w})$.

The minimization of $\alpha$-divergences has also been explored in the context of dropout in \cite{li2017dropout}.
That work considers the same objective as the one optimized by our approach in Section \ref{sec:reparametrization_bba}.
The difference is that the approximate distribution considered by the authors of that work
is limited to the approximate posterior distribution of dropout. This distribution is
given by the mixture of two points of probability mass, \emph{i.e.}, two delta functions,
one of which is located at the origin \cite{gal2016dropout}. The flexibility of this approximate 
distribution is hence very limited. By contrast, the method we propose allows for implicit 
approximate distributions $q(\mathbf{w})$ and therefore is expected to give superior results.

Finally, a closely related method to ours is the one described in \cite{mescheder2017adversarial}. 
This method, Adversarial Variational Bayes (AVB), allows to carry out Variational Inference with
implicit models as the approximate distribution $q(\mathbf{w})$. For this, in that work it is proposed to train a 
discriminator whose output can be used to estimate the KL divergence between the approximate 
distribution $q(\mathbf{w})$ and the prior. This technique has also been considered in other works 
\citep{tran2017hierarchical,huszar2017variational,li2016wild}.
A limitation of AVB is that the method is restricted to minimize the KL divergence between the approximate and the target distribution.
Our approach, by contrast, can optimize the more general $\alpha$-divergence, which includes the KL divergence
as a particular case. Therefore, by changing the $\alpha$ parameter our method can potentially obtain 
better results than AVB. This hypothesis is confirmed by the experiments of the next section.

\section{Experiments}
\label{sec:Experiments}

To analyze and evaluate the performance of the proposed approach, \emph{i.e.}, Adversarial 
$\alpha$-divergence Minimization (AADM), we have carried out extensive experiments, both in synthetic 
data and on common UCI datasets \cite{Dua2017}. Furthermore, we have compared results with previously existing 
methods such as VI, using a factorizing Gaussian as the approximate distribution, and AVB, which 
is a particular case of AADM which optimizes the KL divergence. That is, AADM should give the 
same results as AVB for $\alpha \rightarrow 0$. In these experiments we have also analyzed 
performance versus computational cost of each method on larger datasets with up to 2
million data points. 

The method AADM employed in our experiments consists in the previously described three-network 
system. In particular, the structure we have considered for AADM (and also AVB) is the 
following one: The generator network takes as an input a 100-dimensional Gaussian noise sample,
with adjustable mean and diagonal covariance parameters, and passes 
it through 2 layers of 50 non-linear units each, outputting a sample of the weights $\mathbf{w}$. 
We generate 10 samples for the weights when training, and 50 samples to approximate the 
predictive distribution when testing. Similarly, the discriminator takes these samples of the weights
(as well as samples from the auxiliary distribution described in Section \ref{sec:adaptive_contrast}) 
and passes them through 2 layers of 50 non-linear units each to 
compute $T_\omega(\mathbf{w})$. Finally, the main network (\emph{i.e.}, the model whose weights we are inferring) 
also consists of a 2 layer system with 50 units per layer as well. 
This network uses the sampled weights and the original data as input 
to estimate the AADM objective $\mathcal{L}_\alpha(\phi)$. 
Note that although the network size employed in our experiments is small, 
it is similar to the network size considered in recent related works \cite{hernandez2015probabilistic,li2017dropout}.

The structure described is maintained throughout all the experiments, and remains the same if it is not 
stated otherwise for each specific case. The number of training epochs and the presence (or absence) of a 
warm-up period depends on the dataset being used, and therefore is specified in each experiment.
All non-linear units are leaky RELU units. The code implementing the proposed approach is available
online at \url{https://github.com/simonrsantana/AADM}. All methods have been trained using stochastic optimization
via ADAM \cite{kingma2015}. The learning rate for updating the parameters of the discriminator is set to the default 
value in ADAM, \emph{i.e.}, $10^{-3}$. The learning rate for updating the implicit model 
for $q_\phi$ (\emph{i.e.}, the generator) and the model hyper-parameters (which includes the variance 
of the output noise and the prior) is set to $10^{-4}$. Apart from this, we use the default parameter values in ADAM.
The mini-batch size used is described in each experiment.

\subsection{Synthetic Experiments}

In order to analyze the behavior of the proposed method 
we evaluate the AADM on two simple regression problems  extracted 
from \cite{depeweg2016learning}. More precisely, we generate two different 
\emph{toy datasets}. The first one involving a heteroscedastic predictive 
distribution, and the second one involving a bimodal predictive distribution. 

The structure of the system employed is the one described previously. 
We train this system for 3000 epochs, using the first 500 epochs as the warm-up period. 
We repeat the experiments for different values of alpha in the $(0,1]$. The first dataset 
is generated taking $x$ uniformly distributed in the interval $[-4, 4]$ and $y$ is obtained as $y = 7 \sin x + 3  |\cos(x/2)| \epsilon$, 
where $\epsilon$ is normally-distributed and independent of $x$, \emph{i.e.}, $\epsilon \sim \mathcal{N}(0,1)$. 
Note that this dataset involves input dependent noise.
The second dataset uses $x$ uniformly distributed in the interval$[-2,2]$ and $y = 10 \sin x + \epsilon$ with probability $0.5$ and 
$y = 10 \cos x + \epsilon$ otherwise. 
The distribution of $\epsilon$ is the same as in the first dataset.
note that this other dataset involves a bimodal predictive distribution.
We use 1000 data instances for training and the mini-batch size is set to 10.

\begin{center}
\begin{figure}[H]
  \includegraphics[width = \textwidth] {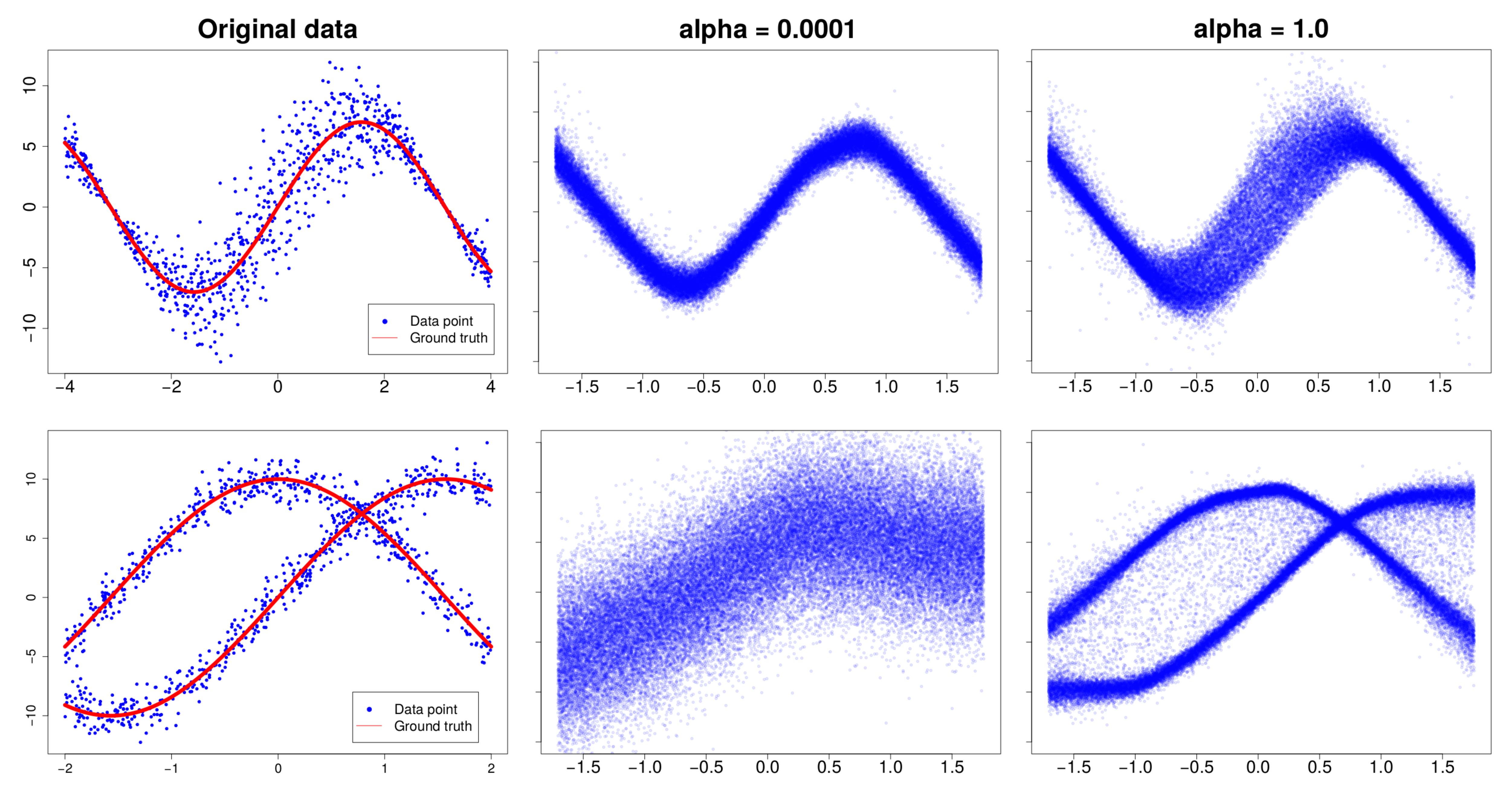}
  \caption{Results for the toy problems. The blue points on the left represent the original 
   training data and the ground truth (red lines). In the middle, predictions generated with 
   $\alpha \approx 0$ (i.e. regular AVB), and in the right side are the predictions with 
	$\alpha = 1.0$.}
  \label{fig:synth_problems}
\end{figure}
\end{center}

The results obtained in the synthetic problems described are represented in Figure \ref{fig:synth_problems}. 
The top figures correspond to the problem involving the heteroscedastic noise and the bottom ones 
to the problem with a bimodal predictive distribution. On the left of the figure we show the original data 
we used to train AADM. In these plots, the red lines represent the \emph{ground truth} for each dataset and 
the blue points are the actual samples we used as training data. The middle and right columns show samples from 
the predictive distribution of a neural network trained using AADM, for $\alpha = 10^{-4}$ and $\alpha = 1$, 
respectively. The results obtained for $\alpha = 10^{-4}$ are expected to be equal to those of AVB.
As can be noticed, a low value of alpha is unable to reproduce the complex structure of the data, losing 
main qualities such as the heteroscedastic additive noise in the first task and the 
bimodality of the predictive distribution in the second task. However, both of them are recovered with accuracy 
when alpha is higher, which is showcased by the results obtained when $\alpha = 1$. 

The results obtained in these experiments, although synthetic, already show that choosing one value of
$\alpha$ or another, for the divergence that is approximately optimized in AADM can significantly change the 
results obtained. In particular, when $\alpha = 10^{-4}$ we can observe that the predictive distribution
that is obtained (after fitting the posterior approximation) focuses more on minimizing the squared error and
less on the log-likelihood of the data. By contrast, when $\alpha=1.0$, the predictive distribution plays
a closer attention to the log-likelihood of the data, and can hence obtain a more accurate predictive distribution.
As can be seen in Table \ref{tab:toy_results}, although the squared error obtained when $\alpha = 10^{-4}$ and $\alpha=1.0$ is very similar,
the test log-likelihood obtained when $\alpha = 1.0$ is much better, which indicates that this value of $\alpha$ produces 
more accurate predictive distributions. Note that the squared error only measures the expected squared deviation from 
target value. The test log-likelihood, on the other hand, measures the overall quality of the predictive distribution,
taking into account, for example, features such as multiple-modes, heavy-tails or skewness.

\begin{table}[]
\caption{Log-likelihood and error results for AADM with $\alpha =10^{-4}$ and $\alpha = 1.0$ in both toy experiments.}
\label{tab:toy_results}
\begin{center}
\begin{tabular}{l|cc|cc|}
\cline{2-5}
 & \multicolumn{2}{c|}{\textbf{Bimodal}} & \multicolumn{2}{c|}{\textbf{Heteroscedastic}} \\ \hline
\multicolumn{1}{|c|}{$\boldsymbol{\alpha}$} & \textbf{Log-likelihood} & \textbf{RMSE} & \textbf{Log-likelihood} & \textbf{RMSE} \\ \hline
\multicolumn{1}{|l|}{$10^{-4}$} & 3.05 & 5.10 & 2.08 & 1.91 \\
\multicolumn{1}{|l|}{$1.0$} & 2.17 & 5.18 & 1.91 & 1.94 \\ \hline
\end{tabular}
\end{center}
\end{table}    

Finally, other values of $\alpha$ give similar results (not shown here). In particular, for
$\alpha < 0.5$ similar results to those of $\alpha = 10^{-4}$ are obtained. By contrast,
when $\alpha > 0.5$ similar results to those of $\alpha = 1.0$ are obtained (that is, only if the training procedure is carried out carefully to avoid bad local optima).

\subsection{Experiments on UCI Datasets}

To analyze in more detail the results of the proposed method, AADM, we have considered eight UCI datasets  \cite{Dua2017}
that are widely spread for regression \cite{hernandez2015probabilistic}.
The characteristics of these datasets are displayed in Table \ref{tab:uci_datasets}.  
Each dataset has a different size, and in order to train the different methods until
convergence we have employed a different number of epochs in each case. The number of epochs selected is presented finally in Table \ref{tab:uci_datasets}. Note that, even though 
there are differences in the epochs employed for training, all of the datasets share the same model 
structure, which is the general one described at the beginning of this section. In all these experiments we
employ the first $10\%$ of the total training epochs for \textit{warming-up} before the KL term is completely 
turned on as in \cite{sonderby2016ladder}. Moreover, the batch size is set to be 10 data points, and 
sampling-wise, we perform 10 samples in the training procedure and 100 for 
testing. We split the datasets in a 90\%-10\% for training/testing. The results reported are 
averages over 20 different random splits of the datasets into training and testing. 

\begin{table}
\caption{Characteristics of the UCI datasets used in the experiments.}
\label{tab:uci_datasets}
\begin{center}
\begin{tabular}{lccc}
\hline
{\bf Dataset} & {\bf Instances} & {\bf Attributes} & {\bf Epochs} \\
\hline
Boston & 506 & 13 & 2000\\
Concrete & 1,030 &  8  &  2000 \\
Energy Efficiency & 768 & 8 &  2000\\
Kin8nm & 8,192 &  8  &  400 \\
Naval & 11,934 & 16 &  400\\
Combined Cycle Power Plant  & 9,568 &  4  &  250\\
Wine & 1,599 & 11 & 2000 \\
Yatch & 308 & 6 & 2000\\
\hline
\end{tabular}
\end{center}
\end{table}

We compare the results of AADM with VI using a factorizing Gaussian as the posterior approximation and with 
regular AVB (which should be the same as our algorithm when $\alpha \rightarrow 0$). For all methods we employ the same two-layered system with 50 units per layer. To make fair comparisons we also 
perform the same warm-up period for both AVB and VI as we use in our method. Therefore only 
after the first 10\% of the total number of epochs, the KL term is completely activated in 
the objective function. 

\begin{figure}[H]
\begin{center}
	\begin{tabular}{cc}
    \includegraphics[width = 0.40\textwidth] {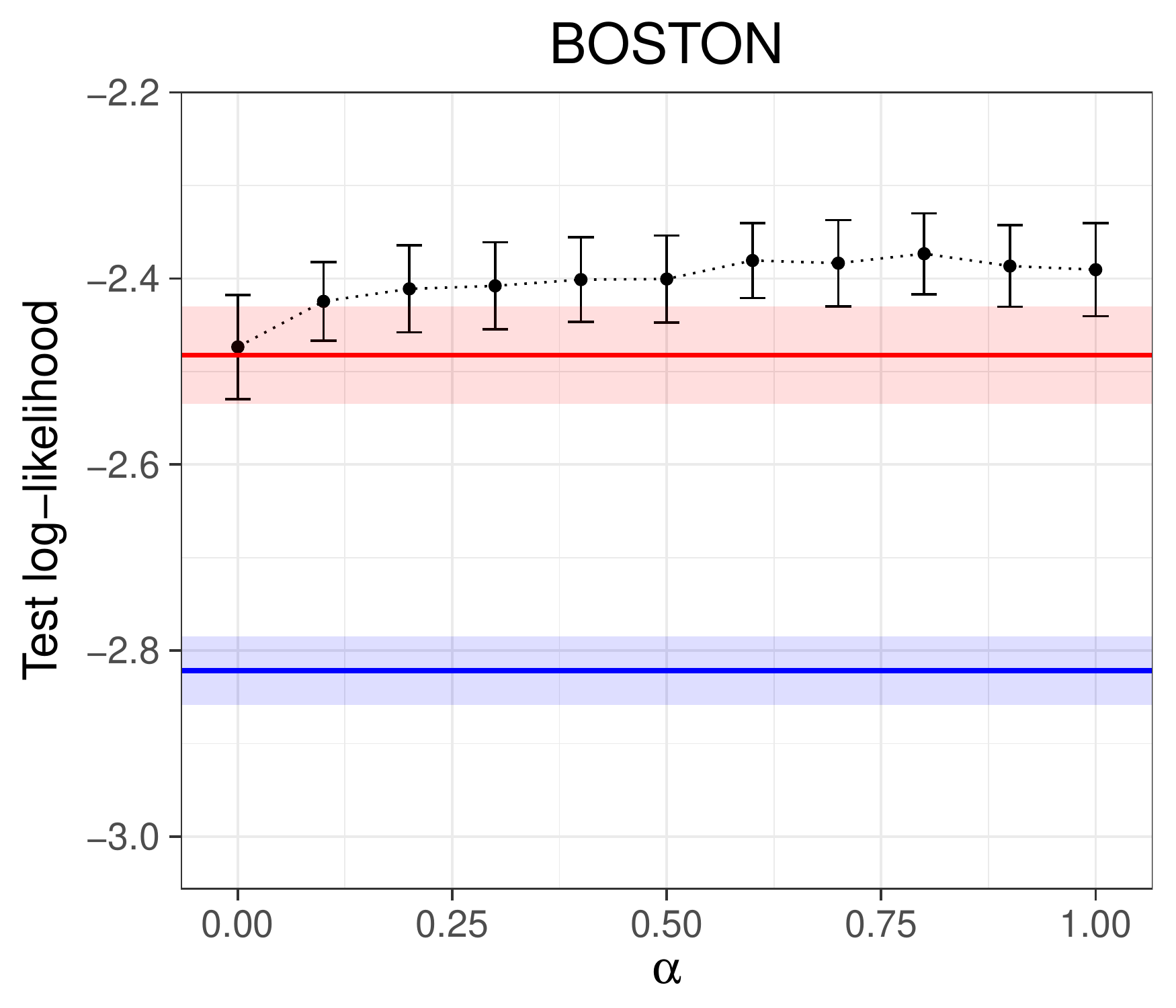} & 
    \includegraphics[width = 0.40\textwidth] {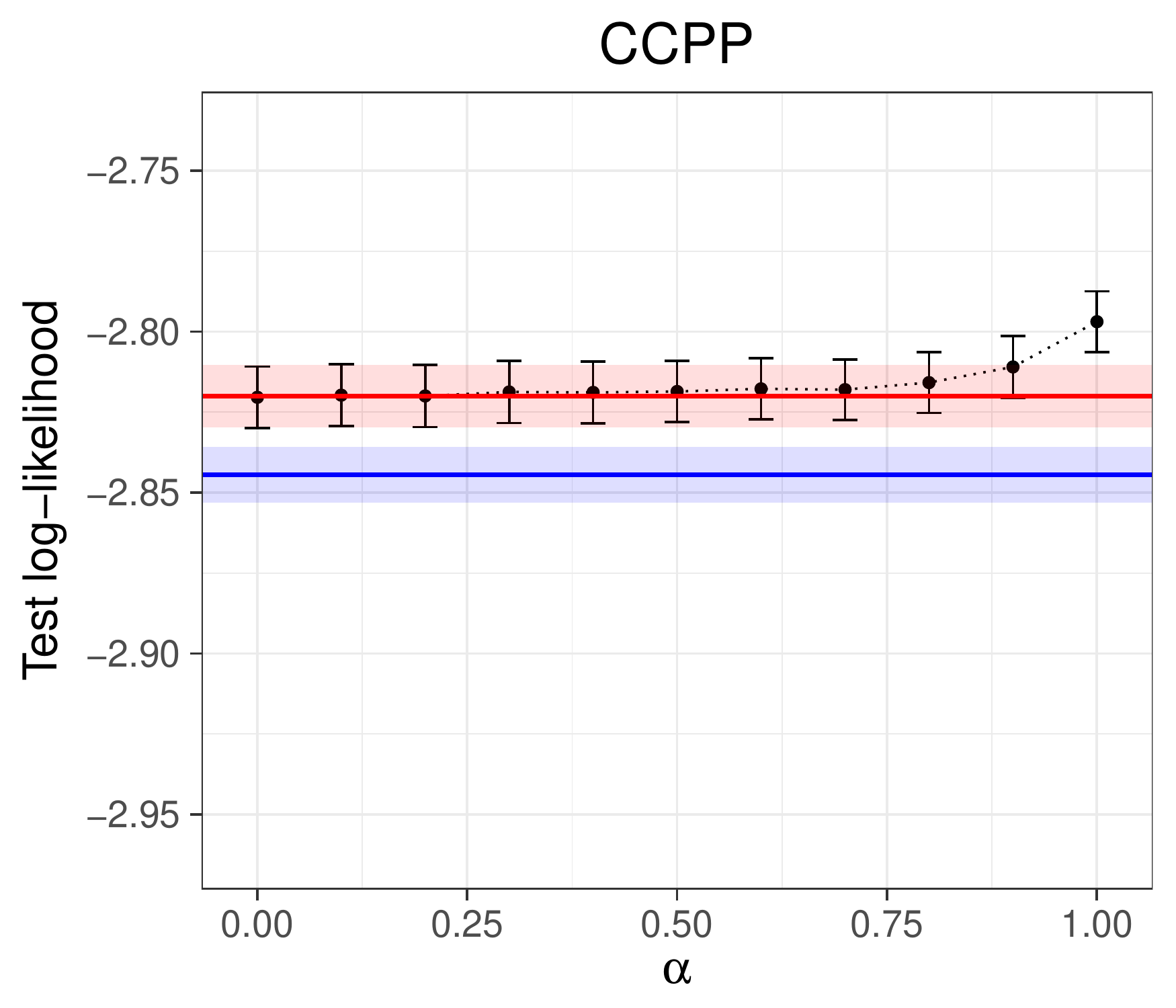} \\
    \includegraphics[width = 0.40\textwidth] {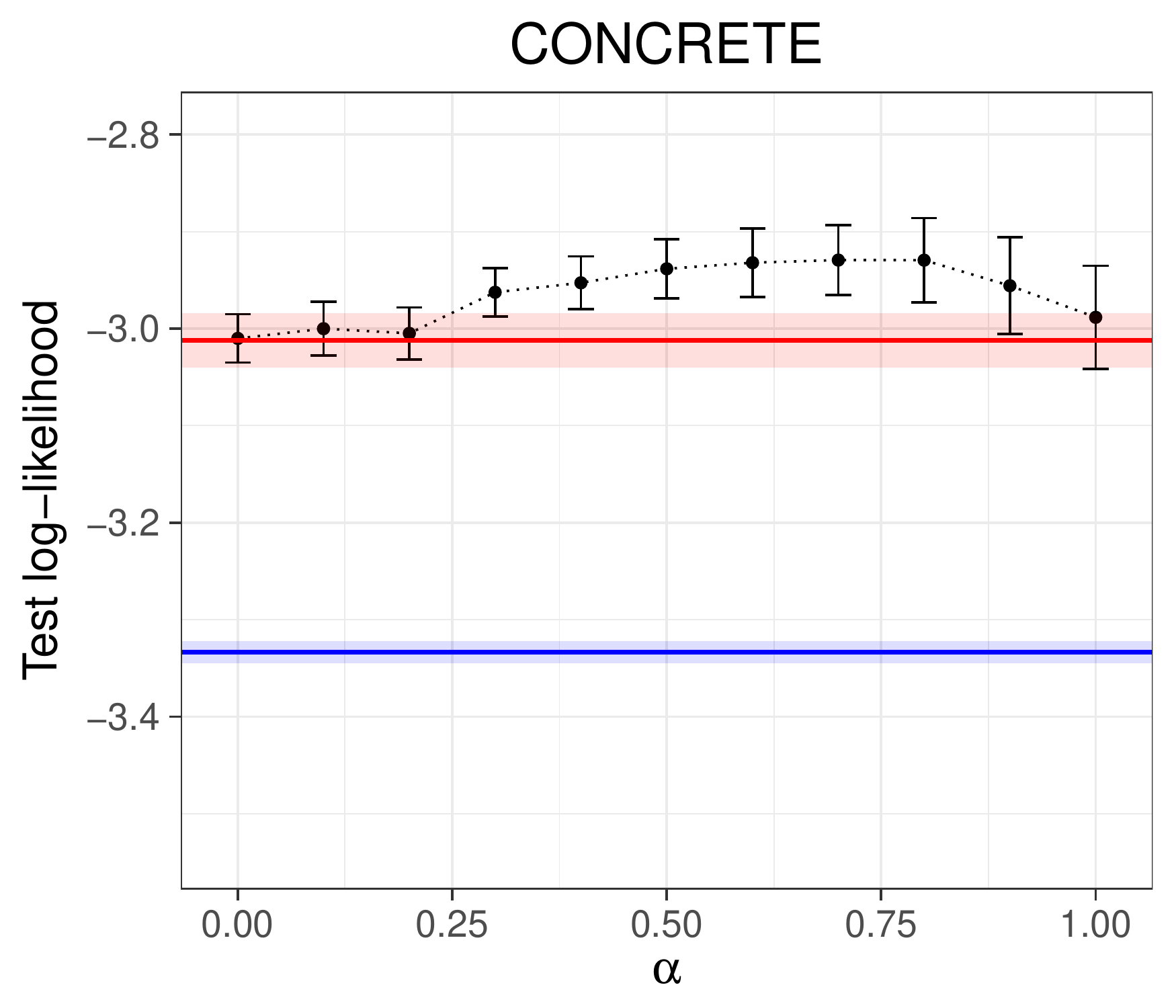} & 
    \includegraphics[width = 0.40\textwidth] {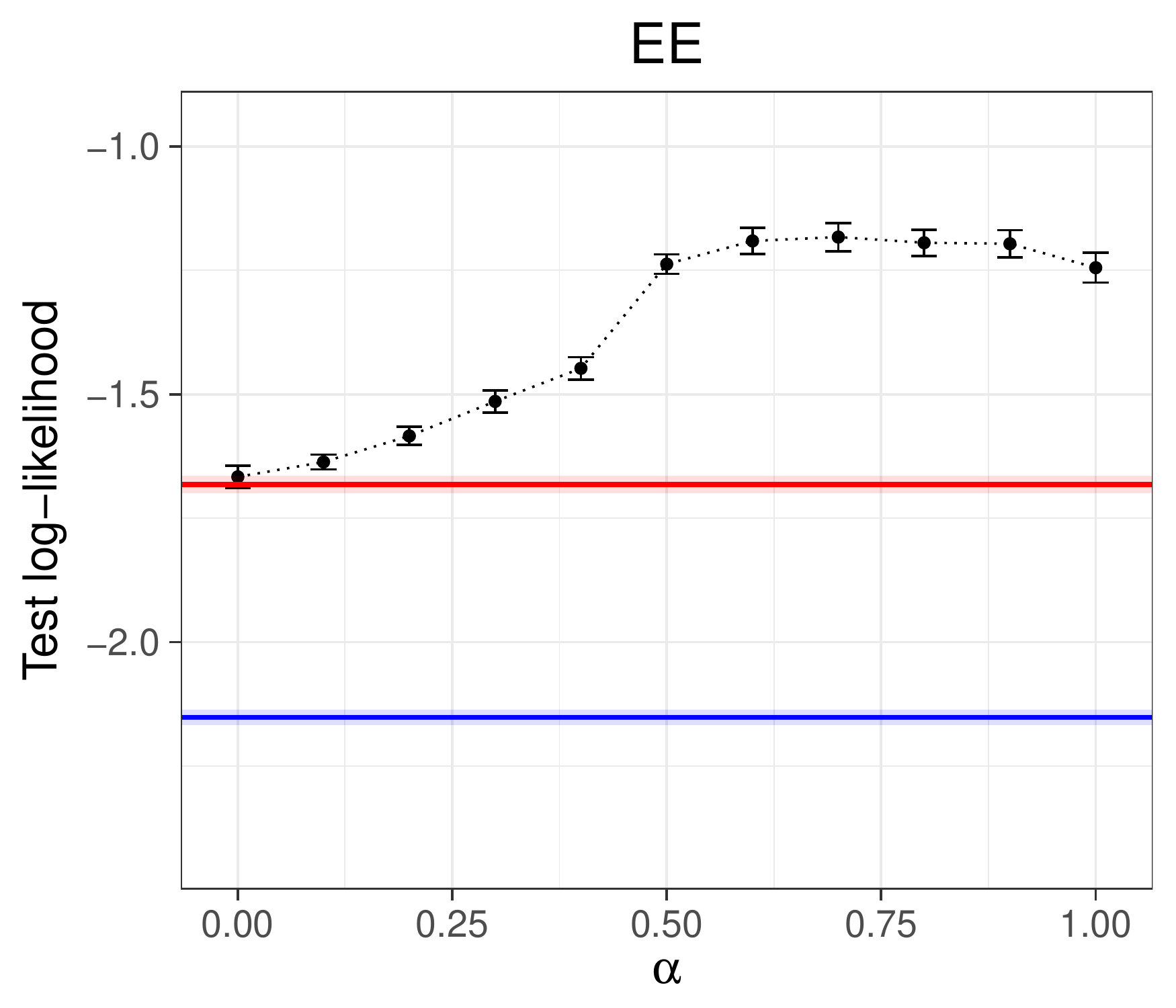} \\
    \includegraphics[width = 0.40\textwidth] {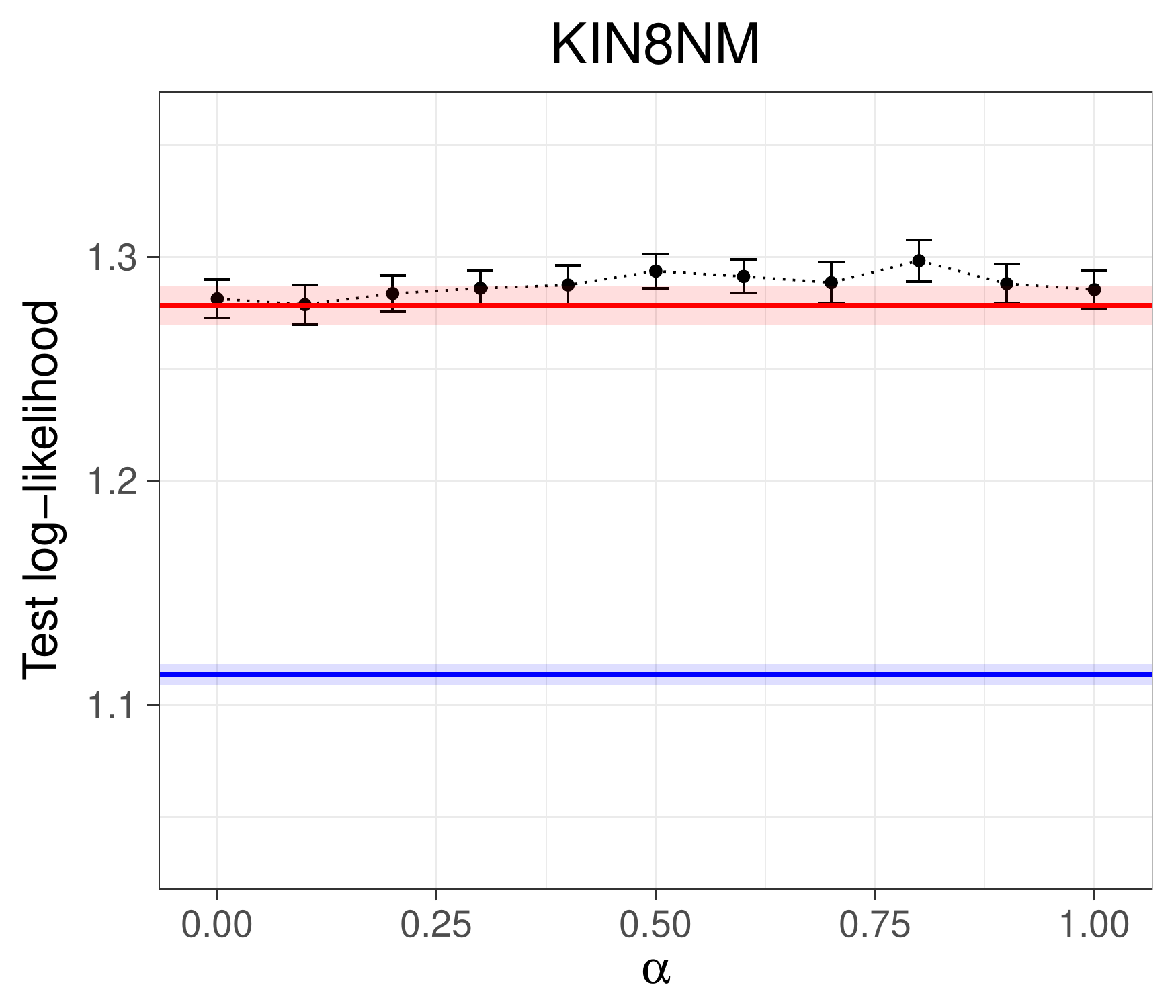} & 
    \includegraphics[width = 0.40\textwidth] {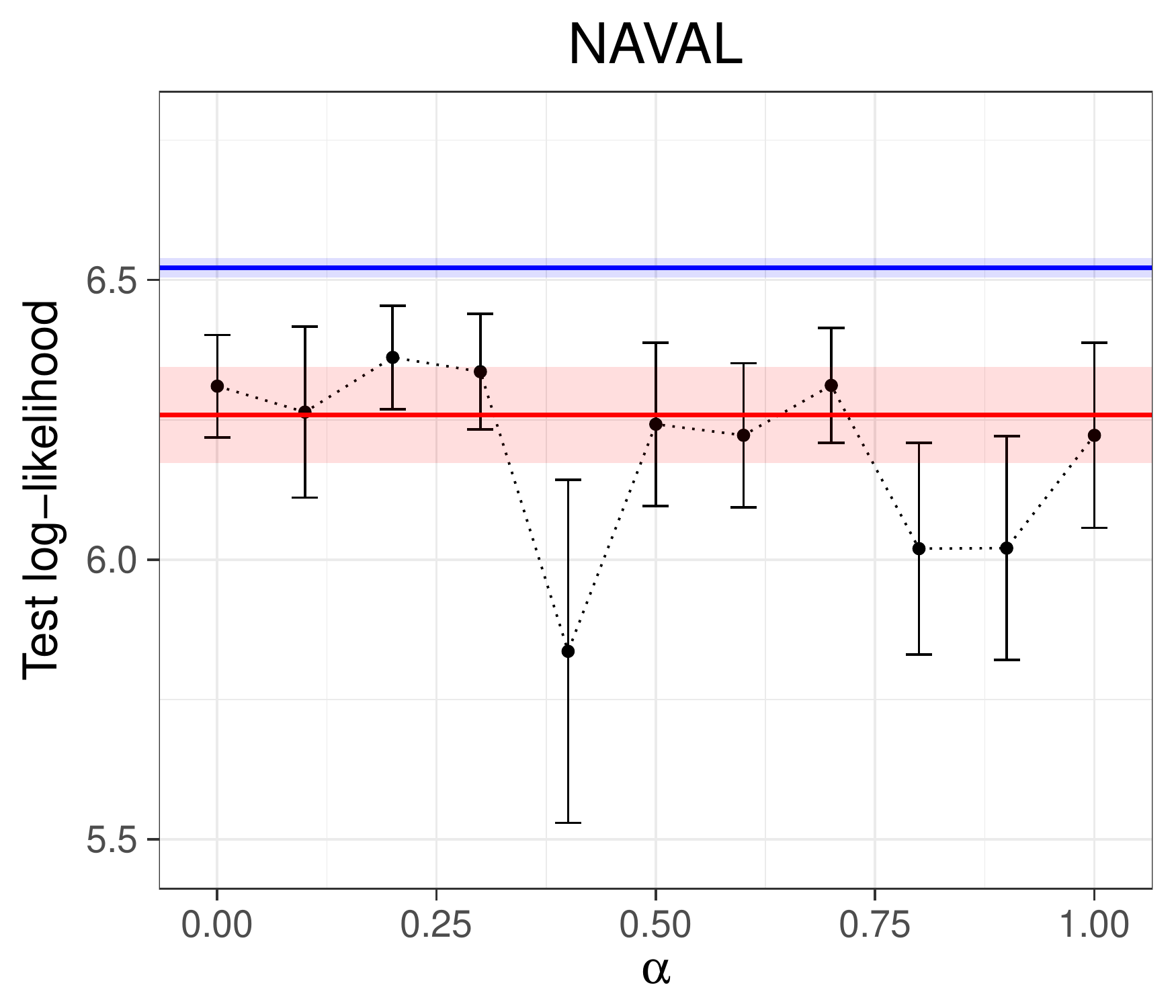} \\
    \includegraphics[width = 0.40\textwidth] {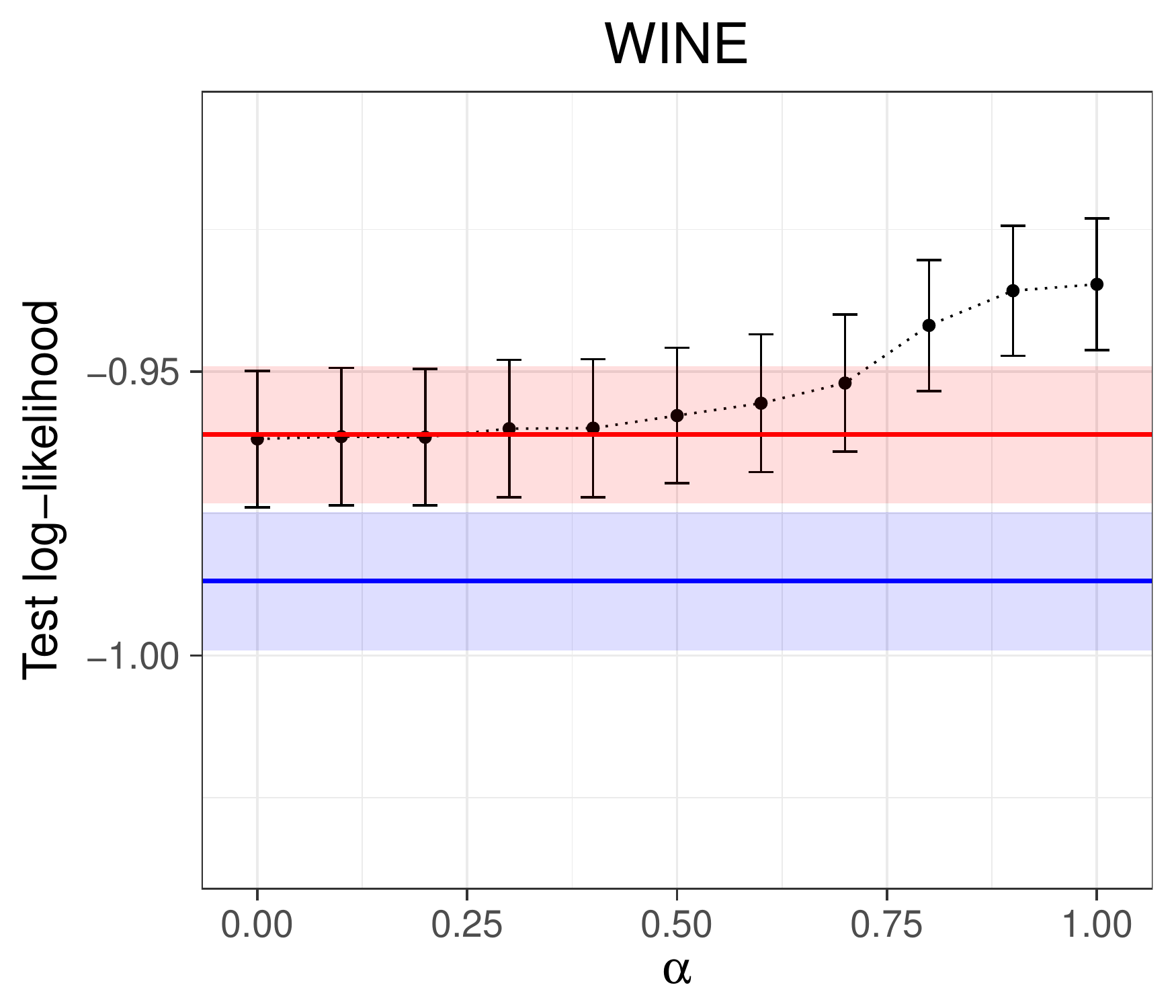} &
    \includegraphics[width = 0.40\textwidth] {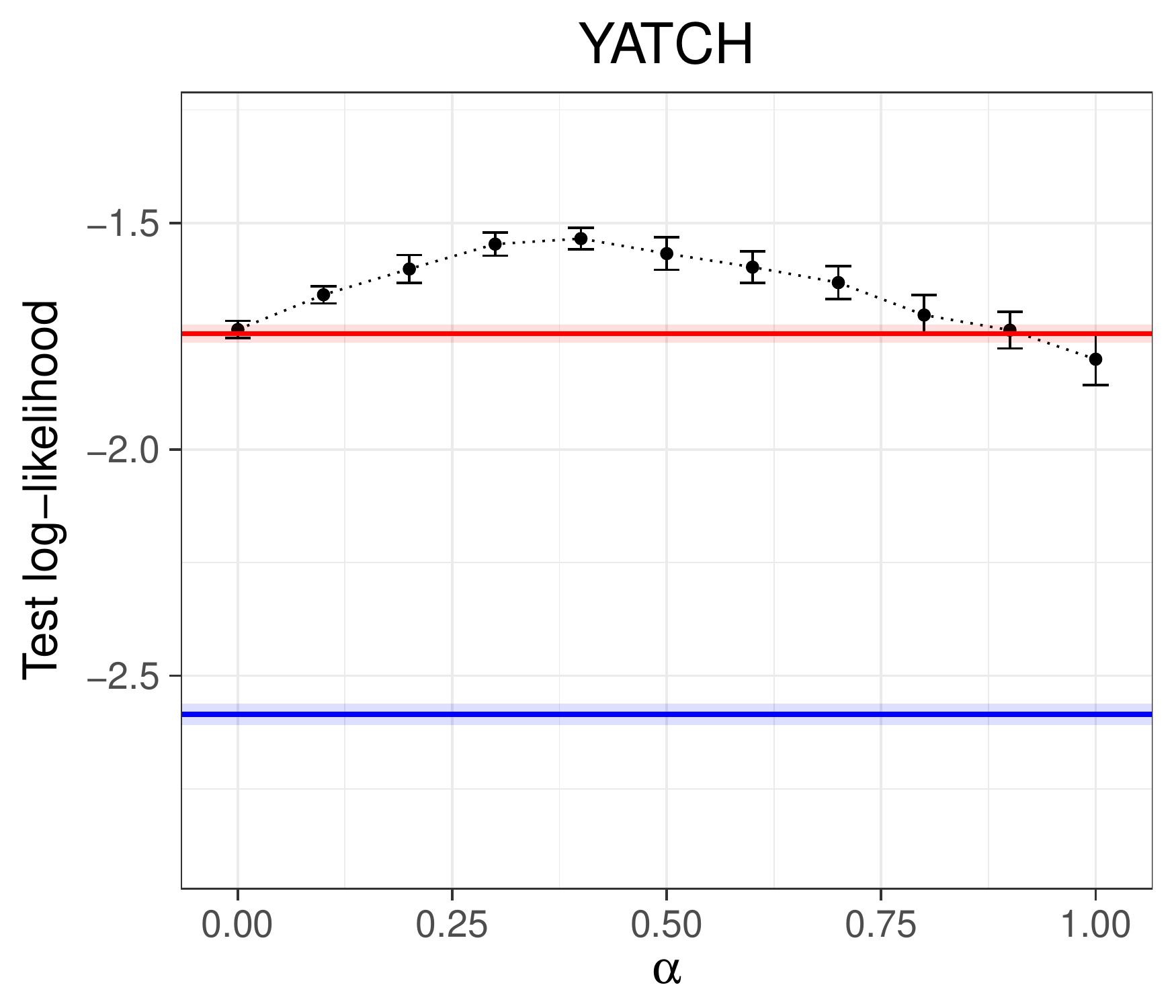} \\
    \multicolumn{2}{c}{\includegraphics[width = 0.35\textwidth] {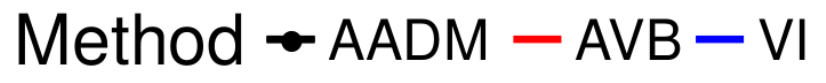}}
	\end{tabular}
  \caption{Average results in terms of the test log-likelihood for the different UCI datasets and methods compared. 
	Black represents the performance for our method, AADM, for different values of $\alpha$. 
	Red is the performance of AVB. VI is presented in blue. Best seen in color.}
  \label{fig:large_figure_experiments_LL}
\end{center}
\end{figure}

The average performance of each method on each dataset, in terms of the test log-likelihood, is displayed Figure \ref{fig:large_figure_experiments_LL}. 
The test log-likelihood measures the overall quality of the predictive distribution,
taking into account, for example, features such as multiple-modes, heavy-tails or skewness.
We observe that values of $\alpha$ that 
are different from $0$ usually outperform both regular AVB and VI in terms of this metric (the higher the values the better).  
From these figures, it seems that  higher values of $\alpha$ often lead to better predictive distributions
it terms of the test log-likelihood, probably as a consequence of being able to better recover  
the real posterior distribution. The values obtained are similar and often better than those of other state of the art methods \cite{hernandez2015probabilistic}. Each of 
the values shown represent the mean performance of a certain method across the 20 different 
splits of each dataset, which are averaged afterwards here. 
Importantly, we observe that standard VI is almost always outperformed by the two techniques 
that allow for implicit models in the posterior approximation $q(\mathbf{w})$. Namely, AVB and AADM. 
This points out the benefits of using an implicit model for the approximate distribution $q(\mathbf{w})$.
Moreover, AVG and AADM give almost the same results when $\alpha \approx 0$, which confirms the 
correctness of our implementation.

\begin{figure}[H]
\begin{center}
	\begin{tabular}{cc}
    \includegraphics[width = 0.40\textwidth]{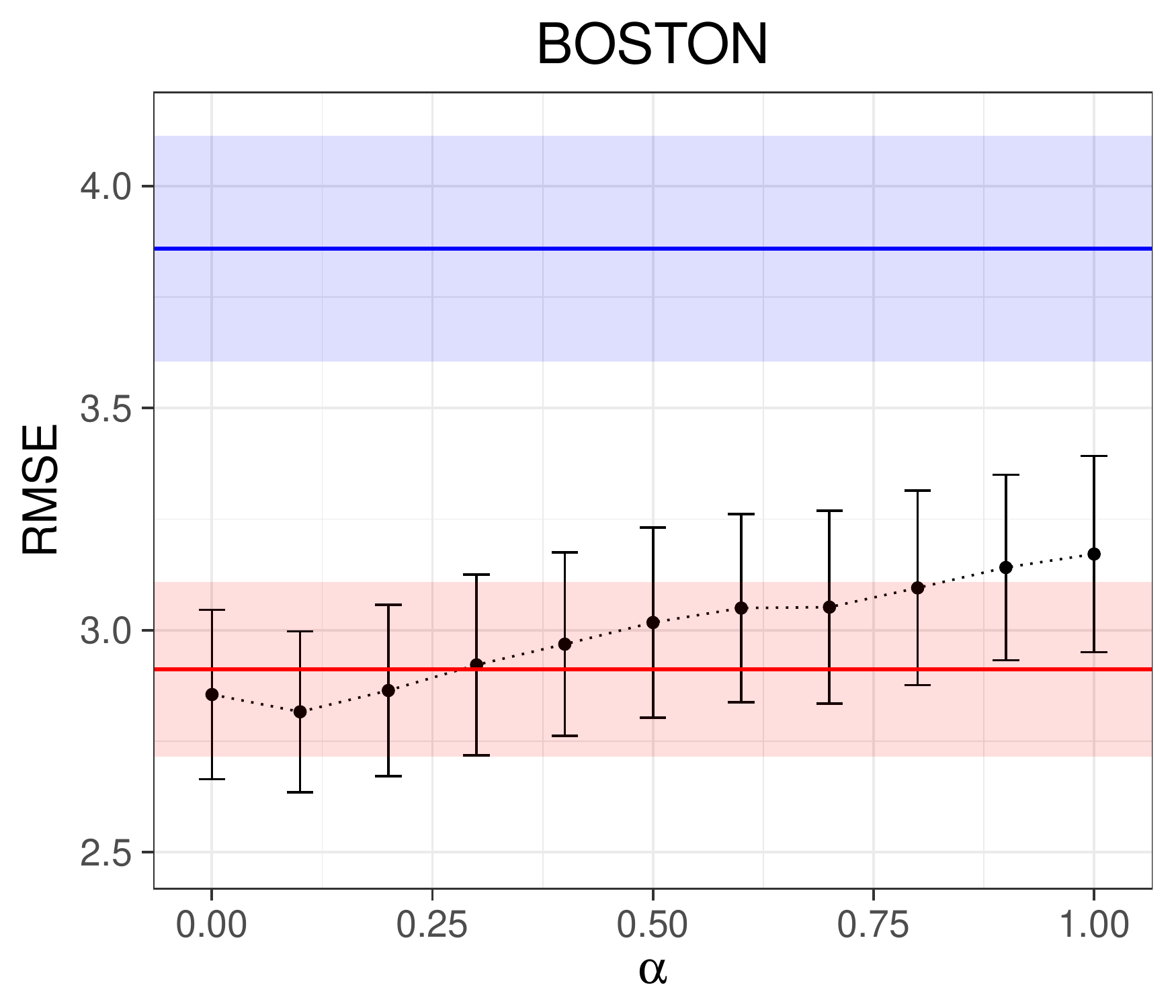} &
    \includegraphics[width = 0.40\textwidth]{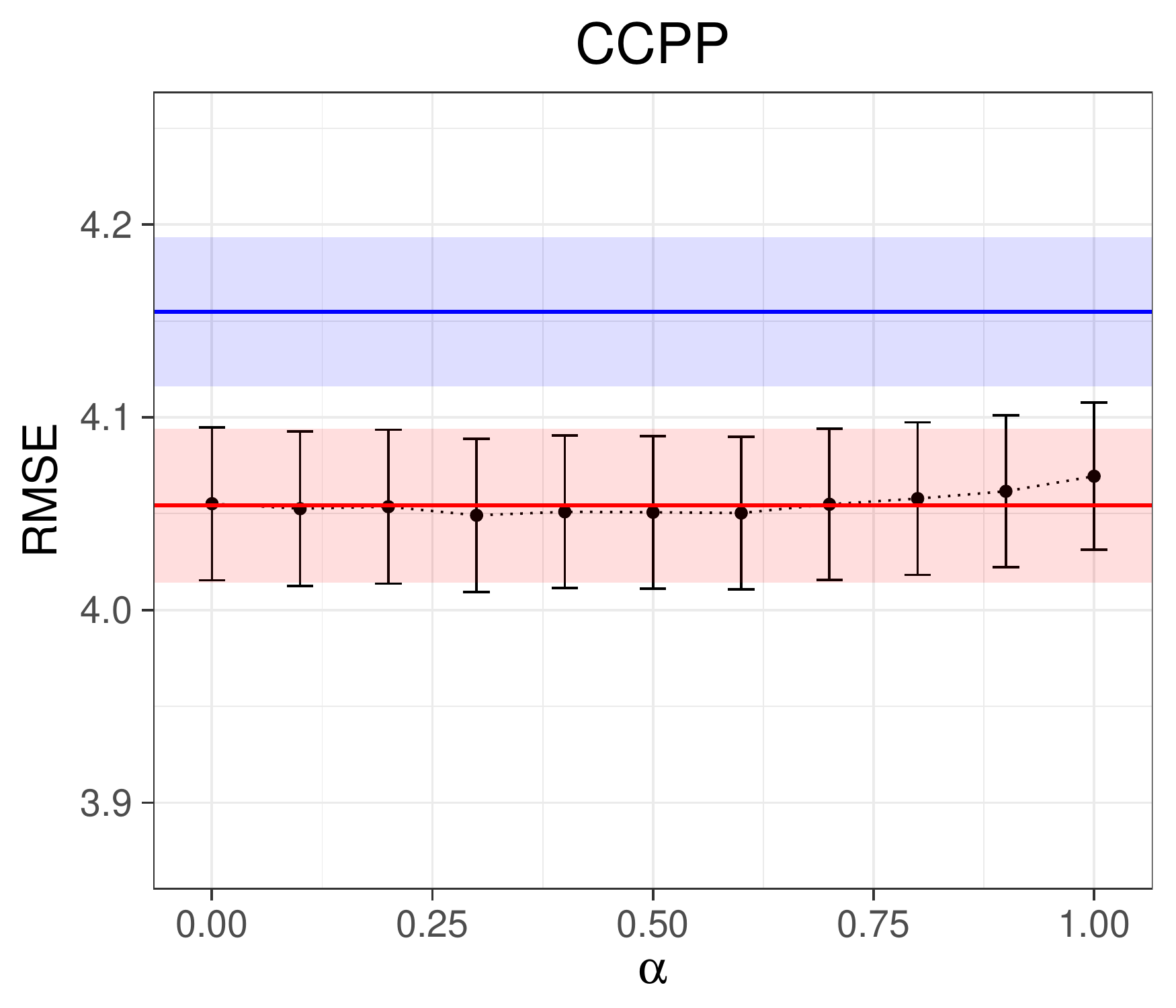} \\
    \includegraphics[width = 0.40\textwidth]{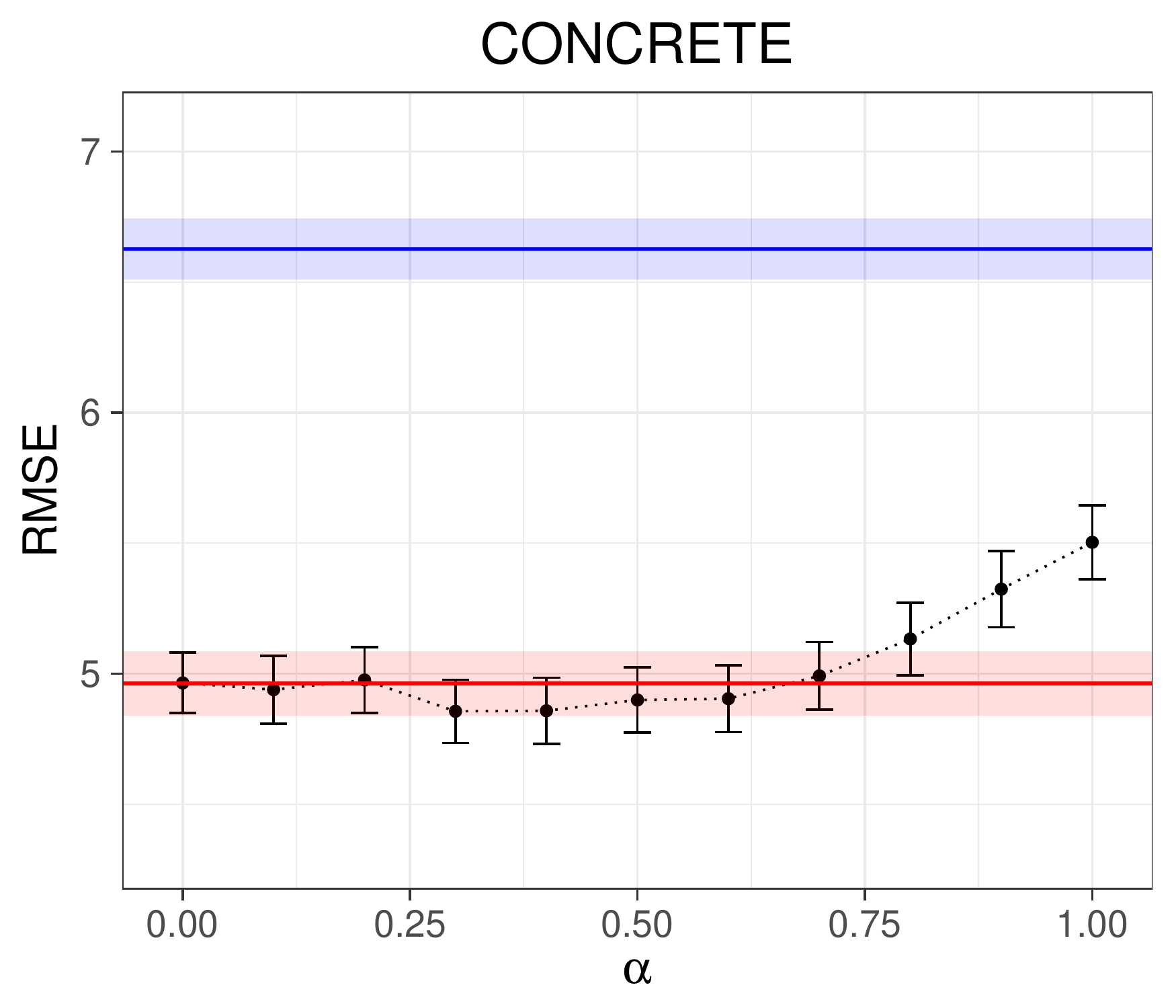} &
    \includegraphics[width = 0.40\textwidth]{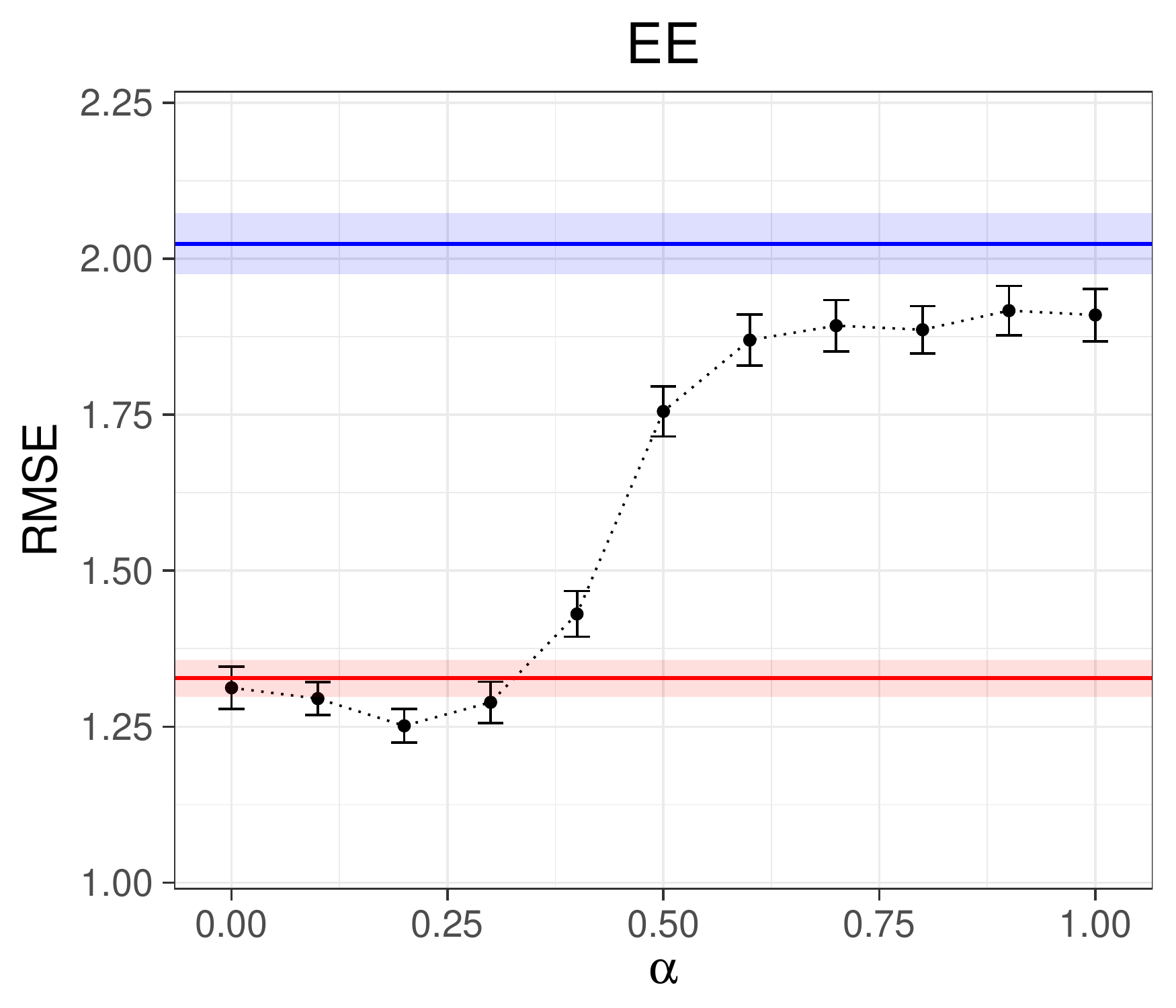} \\
    \includegraphics[width = 0.40\textwidth]{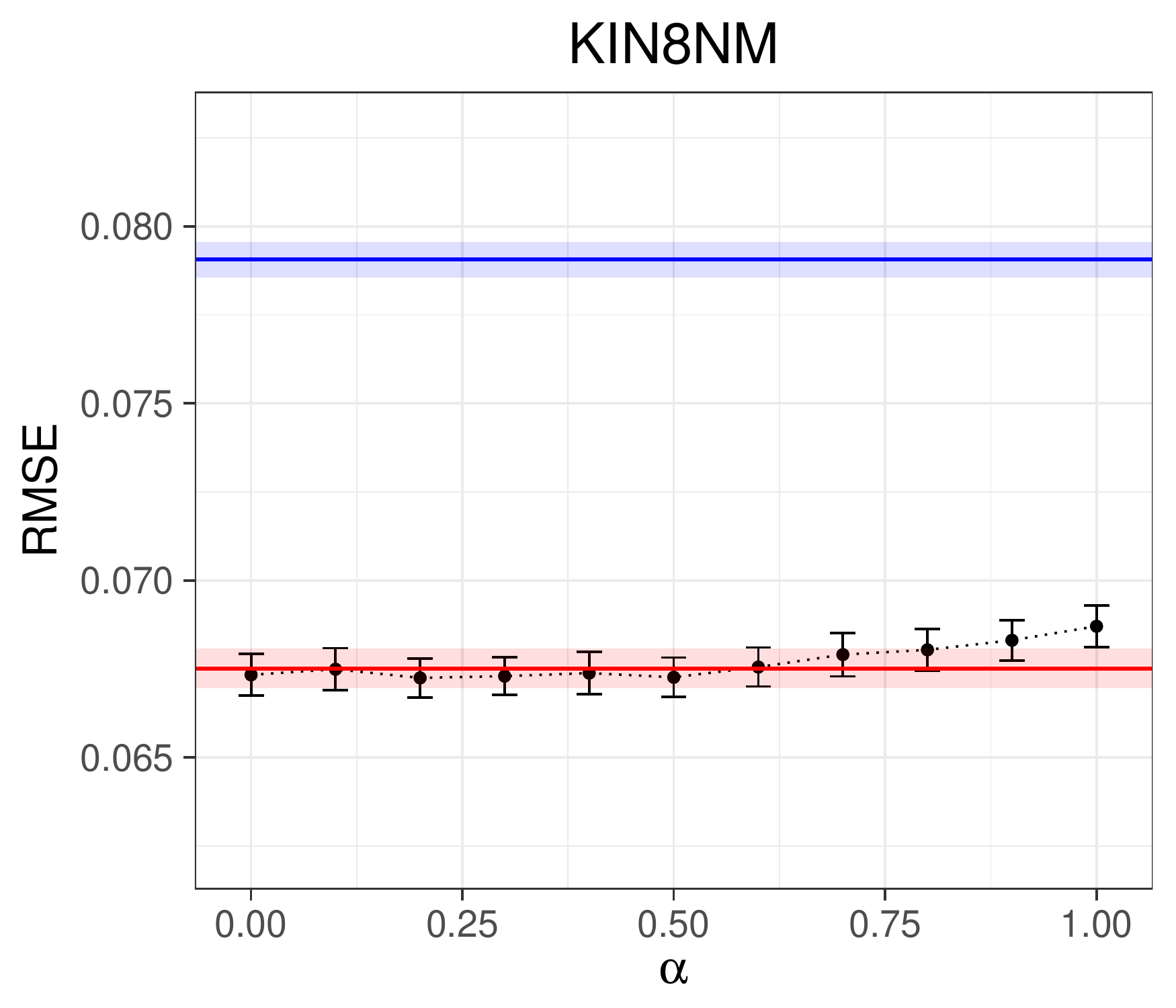} & 
    \includegraphics[width = 0.40\textwidth]{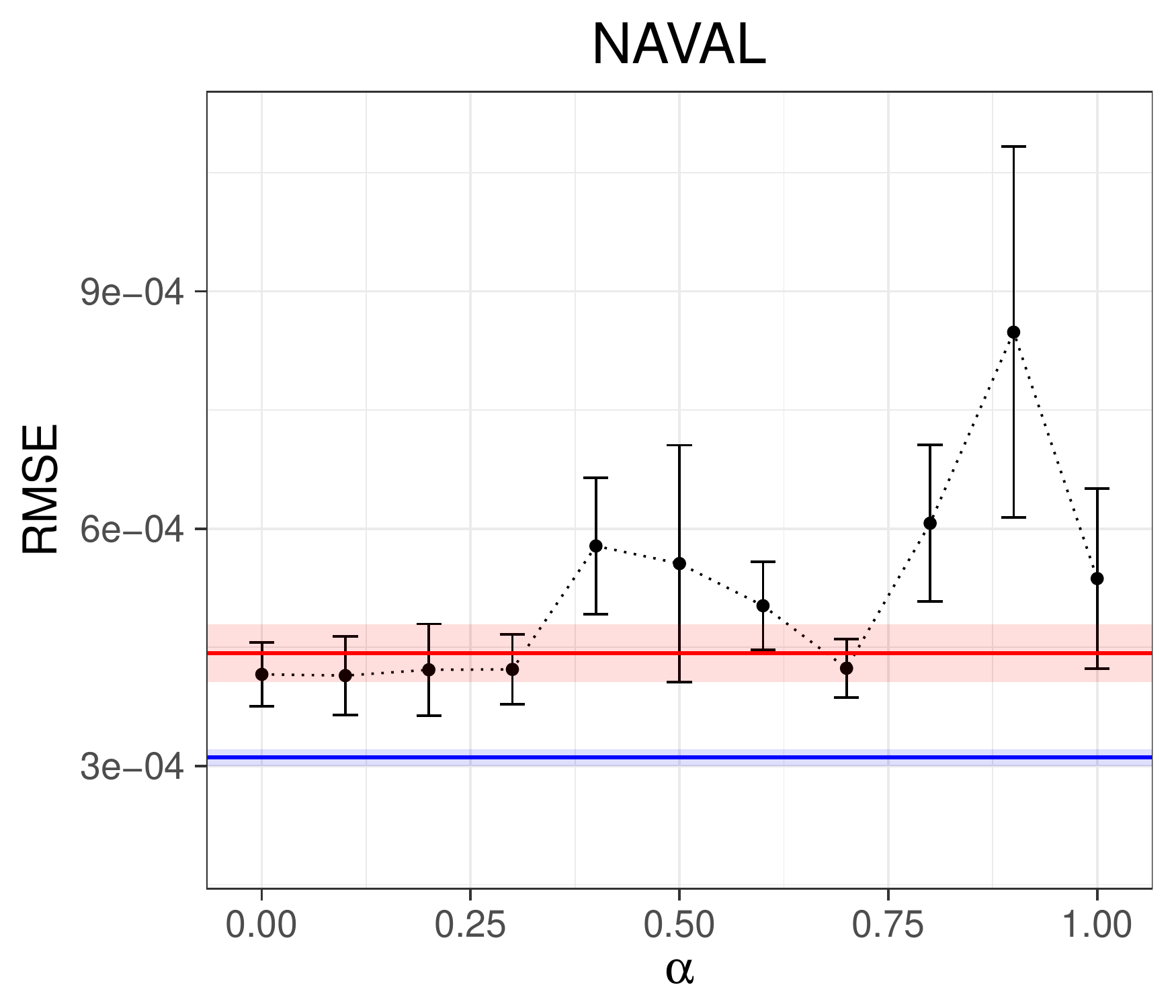} \\
    \includegraphics[width = 0.40\textwidth]{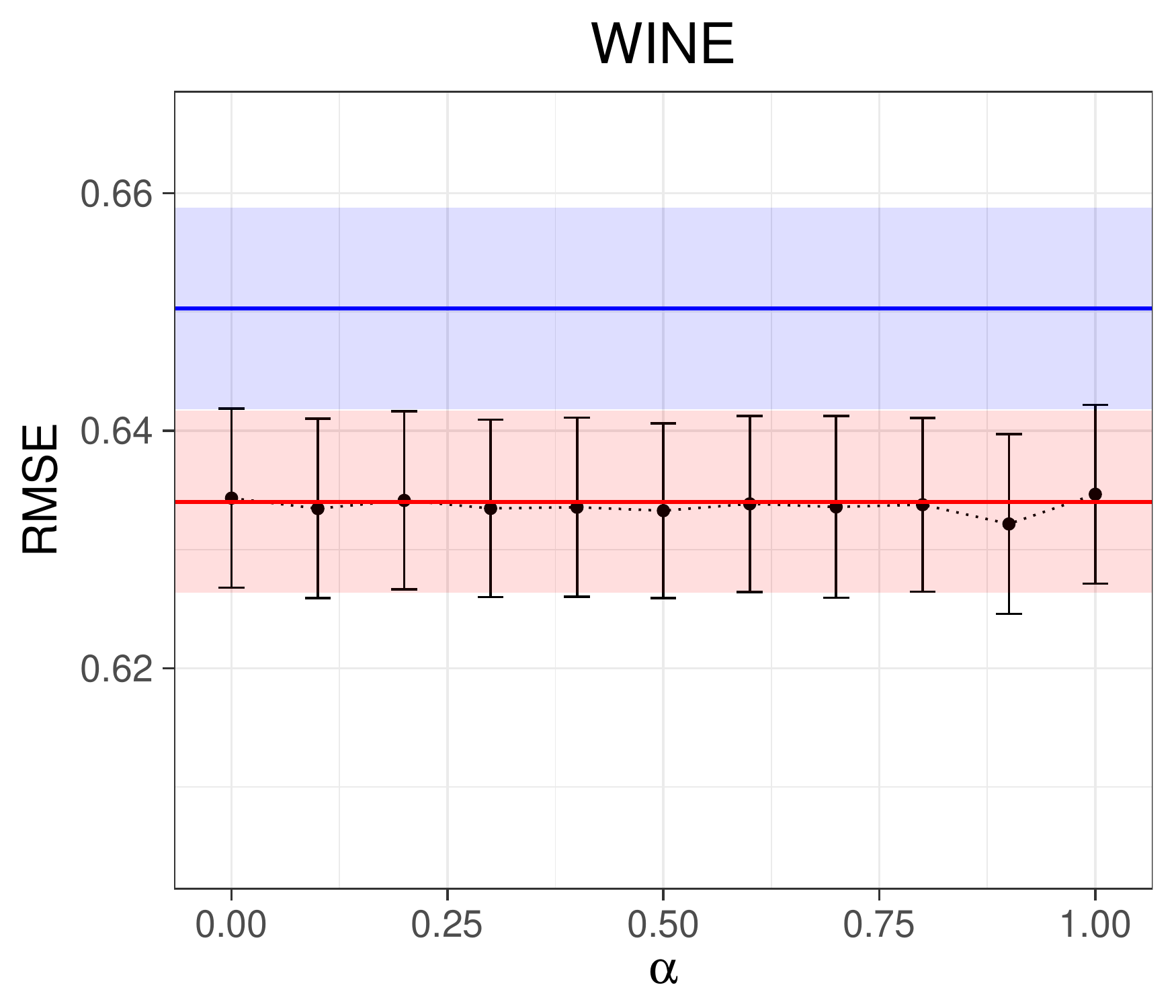} & 
    \includegraphics[width = 0.40\textwidth]{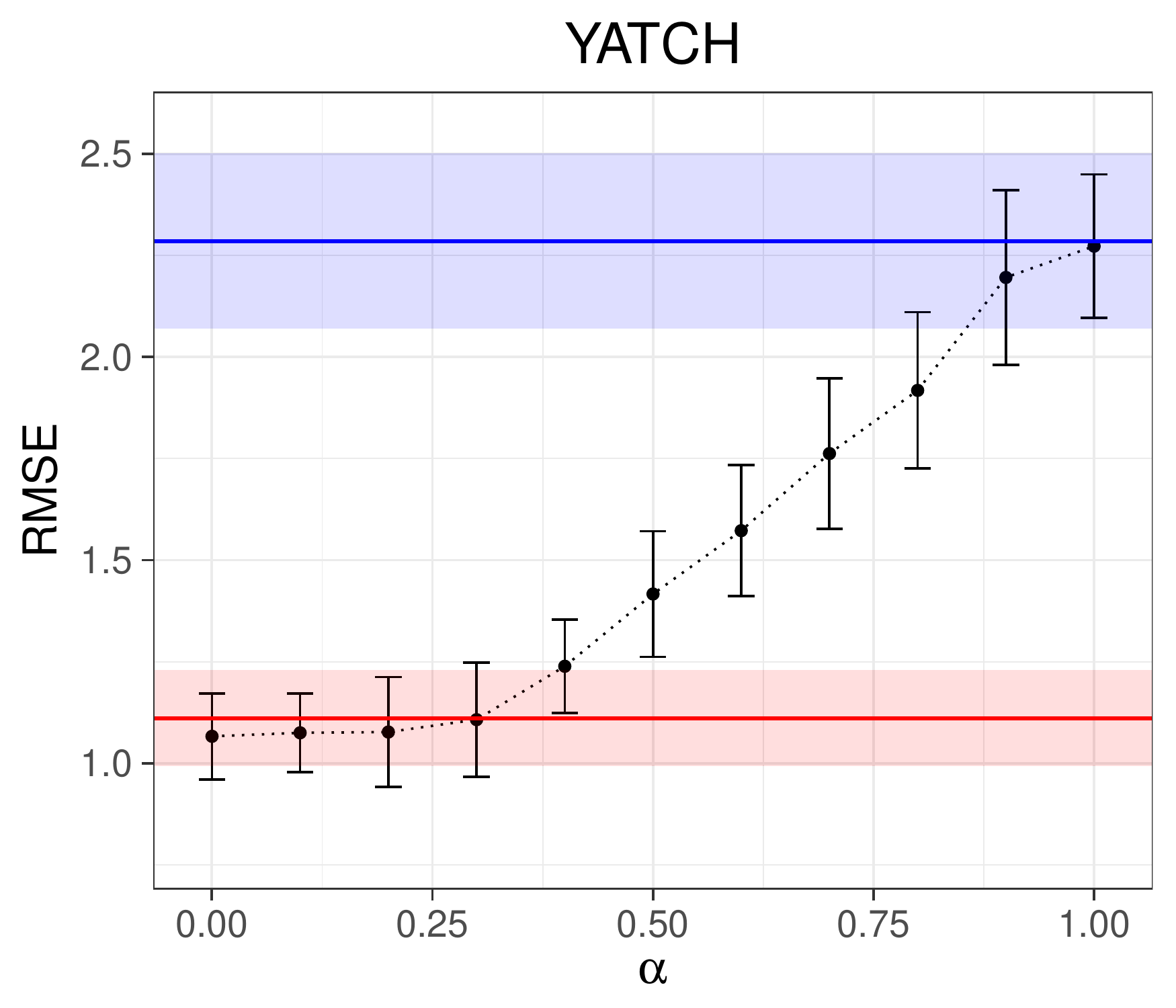} \\
    \multicolumn{2}{c}{\includegraphics[width = 0.35\textwidth] {figures/legend.pdf}}
	\end{tabular}
  \caption{Average results in terms of the root mean squared error for the different UCI datasets and methods compared. 
	Black represents the performance for our method, AADM, for different values of $\alpha$. 
	Red is the performance of AVB. VI is presented in blue. Best seen in color.}
  \label{fig:large_figure_experiments_RMSE}
\end{center}
\end{figure}

The average results obtained for each method on each dataset, in terms of the root mean squared error (RMSE)
are displayed in Figure \ref{fig:large_figure_experiments_RMSE}. Note that the root mean squared error only 
measures the expected deviation from the target value and it may ignore if the model captures accurately the 
distribution of the target value. We can see that the proposed approach, AADM, also obtains better results than VI. 
In this case, nonetheless, different $\alpha$ values do not actually improve much over the basic results of AVB, 
and in general we can see that lower values for $\alpha$ are actually better to obtaining a good performance 
in terms of this metric (here, the lower in the graphs the better the performance). This seems to indicate that
one should choose a value for $\alpha$ that is different, depending on the metric they are most interested in. 
These results are consistent in the sense that, as pointed out previously, values of $\alpha$ close to zero actually
lead to the objective that is optimized in AVB and VI, which pays more attention to the training RMSE, in the
case of regression problems with Gaussian noise. By contrast, values of $\alpha$ closer to one, pay 
more attention to the log-likelihood of the training data.

\subsubsection{Average Rank Results on the UCI Datasets}

\begin{center}
\begin{figure}[H]
    \includegraphics[width = 0.99\textwidth]{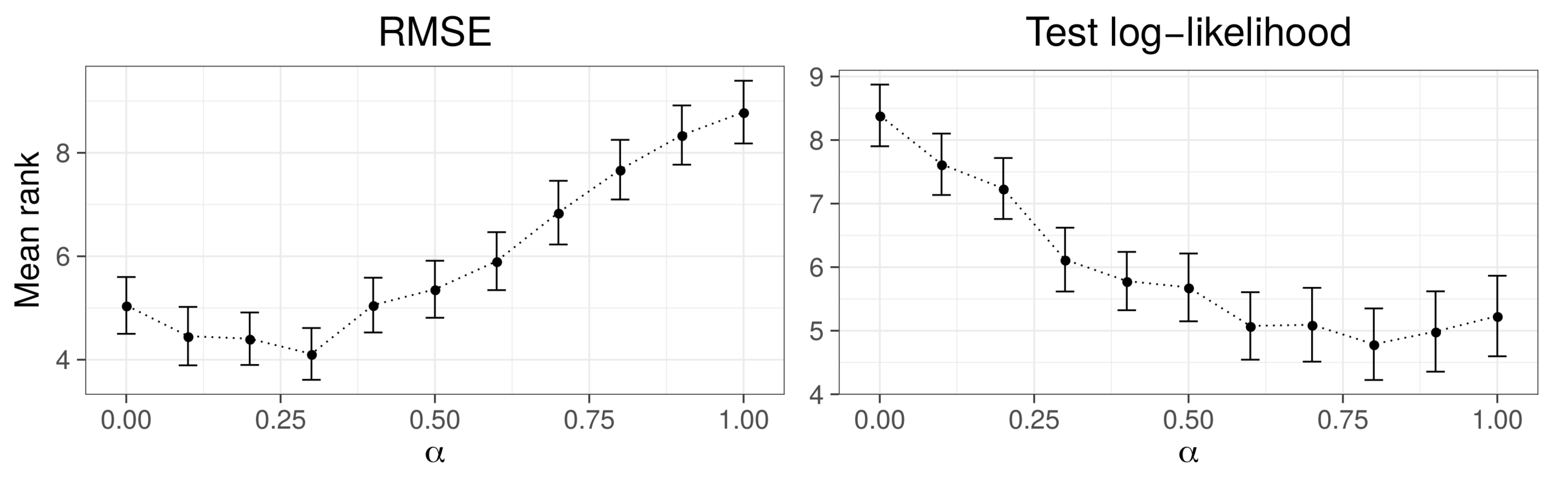}
    \caption{Average rank (the lower the better) for AADM and each value of $\alpha$ in terms of the 
	RMSE (left) and the test log-likelihood (right) across all the UCI datasets and splits.}
    \label{fig:final_rank_analysis}
\end{figure}
\end{center}

To get an overall idea about the performance of AADM, for each value of $\alpha$,
on the previous experiments we have proceeded as follows: We have ranked the performance 
AADM for each $\alpha$ value (\emph{i.e.}, rank 1 means that value of $\alpha$ gives the best result,
rank 2 means that it gives the second best results, etc.). Then, we have computed the average rank 
over all the train / test splits of the datasets, and have calculated the standard deviation in each case. 
Figure \ref{fig:final_rank_analysis} shows the results obtained for the RMSE and test log-likelihood.

The results obtained are displayed in Figure \ref{fig:final_rank_analysis}. 
This figure confirms that medium values for alpha usually present a better performance 
than the extremes (\emph{i.e.}, $\alpha \approx 0$ or $\alpha = 1$), for both the RMSE and the test log-likelihood metrics. 
Furthermore, in the case of the test log-likelihood, higher values of $\alpha$ provide a better 
recovery of the predictive distribution (and hence also the posterior), as indicated by the test log-likelihood.  
In spite of this, lower values of $\alpha$ tend to perform better in terms of the RMSE. 
Again, this behavior can be explained by paying attention to the form the objective 
function optimized in both extremes. The VI objective is recovered when $\alpha \rightarrow 0$.
This objective gives higher importance to the squared errors. By contrast, 
a similar objective function to the one of Expectation Propagation is obtained when $\alpha = 1$. 
This objective includes terms that involve the log-likelihood of the training data.
The main conclusion from this analysis is that the optimal value for $\alpha$ depends on the 
metric we are considering, and that intermediate values of $\alpha$, different 
from $0$ or $1$ are expected to provide the best results.

In order to make a complete analysis on the performance of AADM we have also tested it with 
several binary classification tasks on popular benchmark datasets. The results of this experiments 
can be consulted in the supplementary material provided to this article. In those experiments, AADM 
has shown to be competitive as well, improving the overall performance when compared to VI and obtaining 
similar results to those of AVB. 

\subsection{Experiments on Big Datasets}

To evaluate the performance of the proposed method on large datasets, we have carried out
additional experiments considering two datasets: \emph{Airlines Delay}, and \textit{Year Prediction MSD}.
Airlines Delay  contains information about all commercial flights in the USA from January 2008 to April 2008
\citep{HensmanMG13}. The task of interest is to predict the delay in minutes of a flight 
based on 8 attributes: age of the aircraft, distance that needs to be covered, air-time, departure 
time, arrival time, day of the week, day of the month and month. This is hence a very noisy dataset.
After removing instances with missing values, $2,127,068$ instances remain. 
From these, $10,000$ are used for testing and the rest are used for 
training. \textit{Year Prediction MSD} is publicly accessible on the UCI repository \cite{Dua2017}. 
This dataset has $515,345$ data instances and $90$ attributes. Again, we use $10,000$ for testing and the 
rest of the data are used for training. 
In these experiments the mini-batch size has been set to 100 and we have not used the warm-up annealing 
scheme that deactivates the KL term in the objective of each method during the initial training iterations.
For each method, we measured the performance in the test set, in terms of the RMSE and the test 
log-likelihood, as a function of the training time.

\begin{center}
\begin{figure}[H]
    \includegraphics[width = 0.99\textwidth]{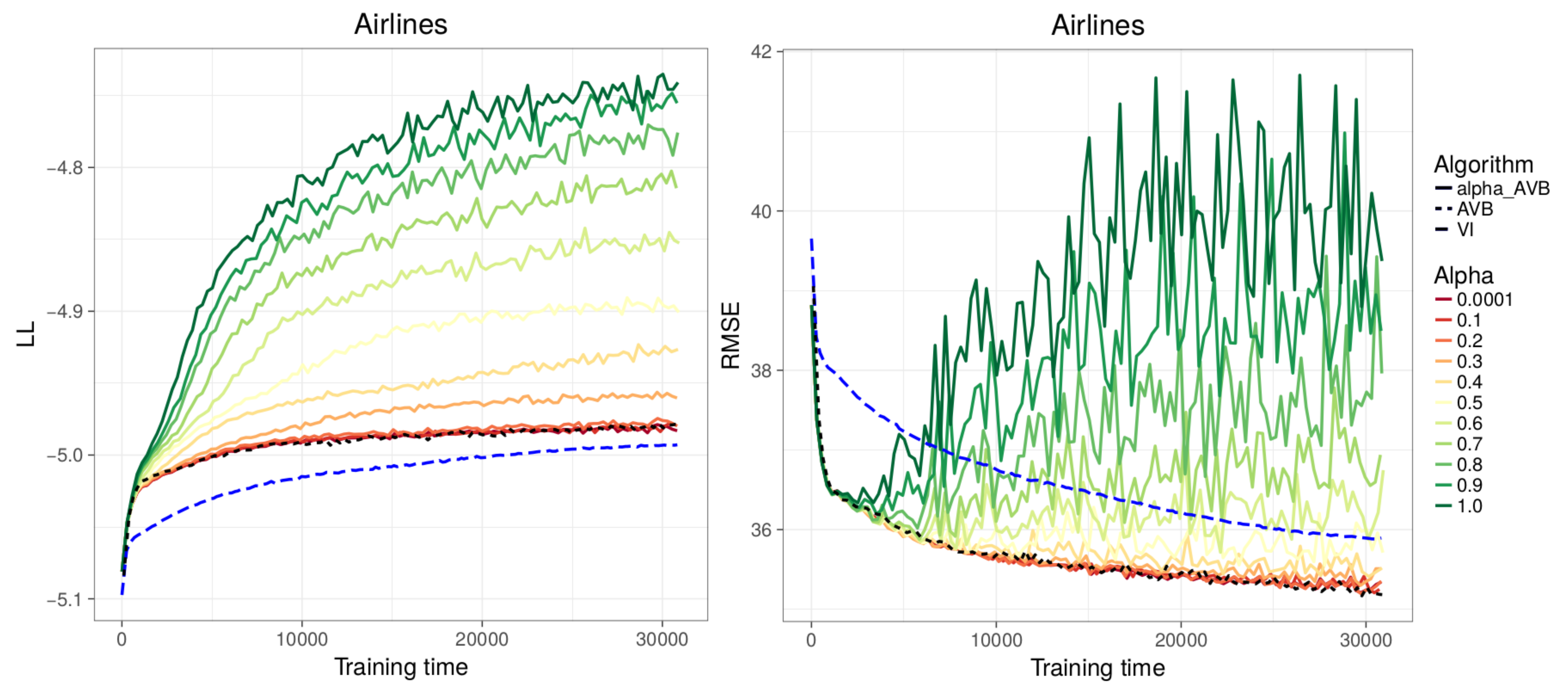}
    \caption{Performance as a function of the computational time in the Airlines dataset for each method.
	We report both in test log-likelihood (left) and the RMSE (right). The dashed blue 
	line corresponds to the method VI, the dashed black 
	line to AVB, and other solid lines represent our method, AADM, for different values of alpha. Best seen in color.}
    \label{fig:airlines_convergence}
\end{figure}
\end{center}

The results obtained for each method on the Airlines dataset are displayed in Figure \ref{fig:airlines_convergence}.
In this figure dashed lines represent other methods, the black being AVB and the blue VI. Solid lines represent 
our method, AADM, for different values of alpha. The figure shows that AADM obtains better results than AVB and 
VI in terms of the test log-likelihood when $\alpha$ approaches $1$. When $\alpha$ is closer to $0$, AADM, gives 
similar results to those of AVB and VI in the long term. The performance of our method w.r.t. the computational time 
is comparable to that of AVB. In terms of RMSE, however, large values of $\alpha$ seem to exhibit a more 
unstable behavior and in general give worse results. This is probably a consequence of this dataset being very noisy.

\begin{center}
\begin{figure}[H]
    \includegraphics[width = \textwidth]{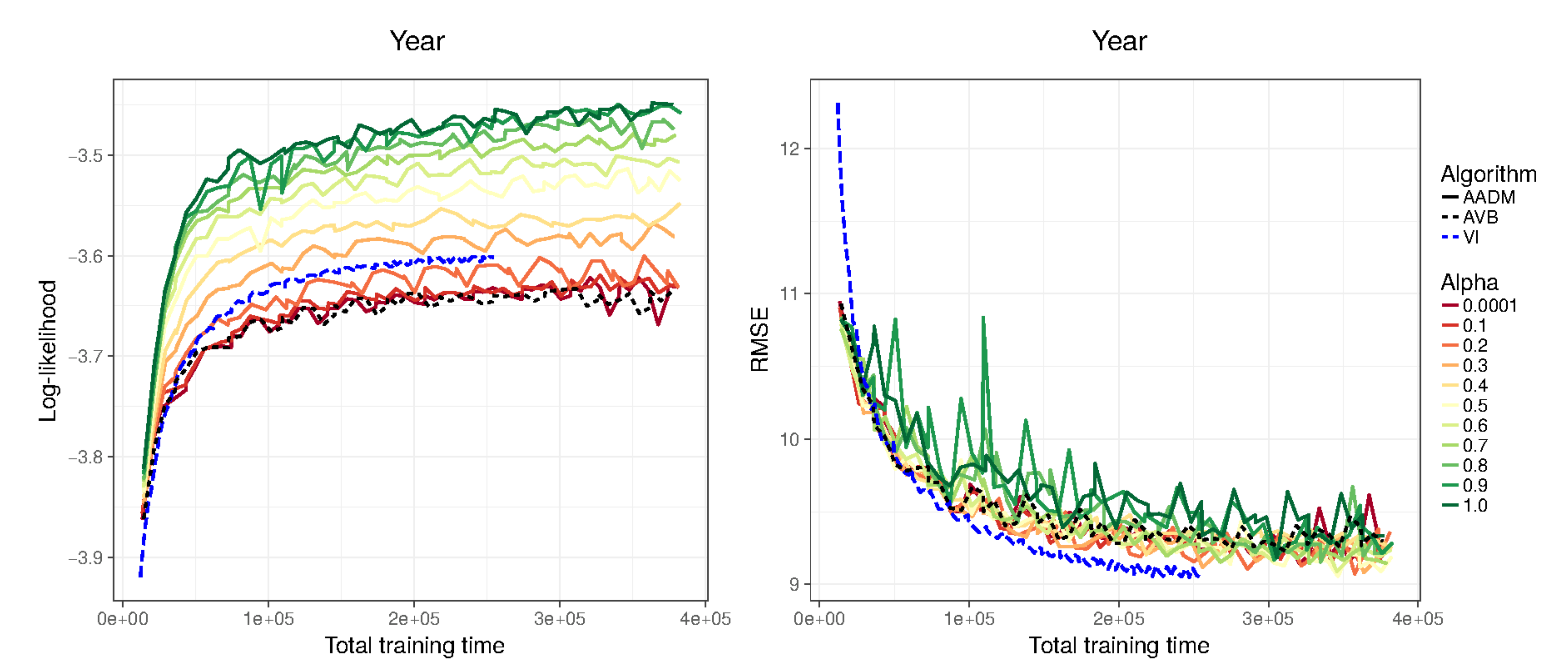}
    \caption{Performance as a function of the computational time in the Year dataset for each method.
	We report both in test log-likelihood (left) and the RMSE (right). The dashed blue 
	line corresponds to the method VI, the dashed black 
	line to AVB, and other solid lines represent our method, AADM, for different values of alpha. Best seen in color.}
    \label{fig:year_convergence}
\end{figure}
\end{center}

The results obtained for each method on the Year dataset are displayed in Figure \ref{fig:year_convergence}.
Again, in this figure dashed lines represent other methods, the black being AVB and the blue VI. Solid lines represent 
our method, AADM, for different values of alpha. As in the previous dataset, AADM obtains better results than AVB and 
VI in terms of the test log-likelihood when $\alpha$ approaches $1$. When $\alpha$ is closer to $0$, AADM, gives 
similar results to those of AVB and VI. In terms of RMSE, lower values of $\alpha$ seems to give also the best
results. However, in this case higher values of $\alpha$ do not seem to give significantly worse results in terms of this
metric.

\section{Conclusions}
 \label{sec:Conclusions}
 
An estimate of the uncertainty in the predictions made by machine learning algorithms
like neural networks is of paramount importance in some specific applications. This estimate can be obtained by following a Bayesian approach. More precisely, the posterior distribution captures which model parameters (neural network weights) are compatible with the observed data. The posterior distribution can then be used to compute a predictive distribution that summarizes the uncertainty in the predictions made. A difficulty, however, is that computing the posterior distribution is intractable and one has to resort to approximate methods in practice.

In this paper we have described a general method for approximate Bayesian inference. The method proposed, named Adversarial $\alpha$-divergence Minimization (AADM), allows to tune an approximate posterior distribution by approximately minimizing the $\alpha$-divergence between this distribution and the posterior. The $\alpha$-divergence generalizes the KL divergence, commonly used to perform approximate inference. AADM also allows to account for implicit models in the approximate posterior distribution. Implicit models allow to specify a probability distribution simply as some non-linear transformation of random input noise. If the non-linear transformation is complex enough, this will lead to a flexible model that is able to represent arbitrarily complex posterior distributions. A drawback of implicit models is, however, that one cannot evaluate the p.d.f. of the resulting distribution, which is required for approximate inference. We overcome this problem by following the approach of \cite{mescheder2017adversarial}, and more precisely, we learn a discriminative model that estimates the log-ratio between the p.d.f. of the implicit model and a much simpler distribution (\emph{i.e.}, a Gaussian distribution).

The proposed method, has been evaluated on several experiments and compared to other methods for approximate inference such as Variational Inference (VI) with a factorizing Gaussian as the approximate distribution, and Adversarial Variational Bayes (AVB) \cite{mescheder2017adversarial}. The experiments carried out, involving approximate inference with Bayesian neural networks, indicate that implicit models almost always provide better results than a factorizing Gaussian in terms of the metrics employed. Moreover, the minimization of $\alpha$-divergences seems to provide overall better results in regression tasks than the plain minimization of the KL divergence, as done by VI and AVB. In particular, values of $\alpha$ that are close, but not exactly equal to $1$ seem to provide better predictive distributions in terms of the test log-likelihood. By contrast, in terms of the root mean squared error (RMSE) one should choose values of $\alpha$ that are close to, but not exactly equal to zero. Therefore, we conclude that one can obtain better results in terms of the test log-likelihood and the RMSE by employing the proposed method, AADM, and by choosing a value of $\alpha$ that may depend on the specific performance metric we are interested in. Future work on this topic may include a detailed analysis on which values of $\alpha$ are optimal depending on the characteristics of the dataset as well as the task at hand, understanding how this optimal values change depending on the size and complexity of the dataset, alongside other similar relevant features.

\section*{Acknowledgements}

Sim\'on Rodr\'iguez acknowledges the Spanish Ministry of Economy for the FPI SEV-2015-0554-16-4 Ph.D. scholarship. 
The authors gratefully acknowledge the use of the facilities of Centro
de Computaci\'on Cient\'ifica (CCC) at Universidad Aut\'onoma de
Madrid. Daniel Hern\'andez-Lobato also acknowledges financial support from Spanish
Plan Nacional I+D+i, grants TIN2016-76406-P and TEC2016-81900-REDT.

\section*{References}

\bibliography{main_bibfile}

\end{document}


\begin{frontmatter}

\title{ \textbf{\textit{Supplementary material}} \\  Adversarial $\alpha$-divergence Minimization for Bayesian Approximate Inference}


\author[first_address]{Sim\'on Rodr\'iguez Santana \corref{mycorrespondingauthor}}
\cortext[mycorrespondingauthor]{Corresponding author}
\ead{simon.rodriguez@icmat.es}

\author[second_address]{Daniel Hern\'andez-Lobato}
\ead{daniel.hernandez@uam.es}

\address[first_address]{Institute of Mathematical Sciences (ICMAT-CSIC), Campus de Cantoblanco, C/Nicol\'as Cabrera, 13-15, 28049 Madrid, Spain.}
\address[second_address]{Escuela Politécnica Superior, Universidad Autónoma de Madrid, Campus de Cantoblanco, C/Franciso Tom\'as y Valiente 11, 28049 Madrid, Spain.}

\end{frontmatter}

\section{Classification experiments}

In order to characterize AADM in a more complete manner we have also conducted experiments on binary classification. We have employed six different datasets which are very common for benchmarking \cite{bui2017unifying, Dua2017}. With all of these datasets we have performed the same analysis done before in regression: we have defined 20 splits for each one and afterwards we have analyzed the average results both in terms of the log-likelihood and the classification error rate for across splits. For each dataset we have trained the methods for 3000 epochs, ensuring convergence. We have also split them in a 90\%-10\% for training and testing, as we did for the regression experiments. All of the datasets share the same model structure, which is the general one described in the main article. In all these experiments we employ the first $10\%$ of the total training epochs for \textit{warming-up} before the KL term is completely turned on as in \cite{sonderby2016ladder} (see the description of the \textit{annealing factor} in the main article for more details into this). Moreover, the batch size is set to be 10 data points, and sampling-wise, we perform 10 samples in the training procedure and 100 for testing. For more precise information on the datasets employed see Table \ref{tab:class_datasets}.

As was done for regression, we compare the results of AADM with VI using a factorizing Gaussian as the posterior approximation and with regular AVB (whose results should be similar to the ones obtained with AADM when $\alpha \rightarrow 0$). To make fair comparisons we also perform the same warm-up period for both AVB and VI as we use in our method. 

\begin{table}[]
\begin{center}
\begin{tabular}{lccc}
\hline
{\bf Dataset} & {\bf N for train/test } & {\bf Attributes} & {\bf Positive/negative instances} \\
\hline
Australian & 621/69 & 15 & 222/468\\
Breast & 614/68 &  11  &  239/443 \\
Crabs & 180/20 & 7 &  100/100\\
Iono & 315/35 &  35  &  126/224 \\
Pima & 690/77 & 9 &  500/267\\
Sonar  & 186/21 &  61  &  111/96\\
\hline
\end{tabular}
\end{center}
\caption{Characteristics of the UCI datasets used in the experiments.}
\label{tab:class_datasets}
\end{table}

\begin{figure}[]
\begin{center}
	\begin{tabular}{ccc}
    \includegraphics[width = 0.31\textwidth] {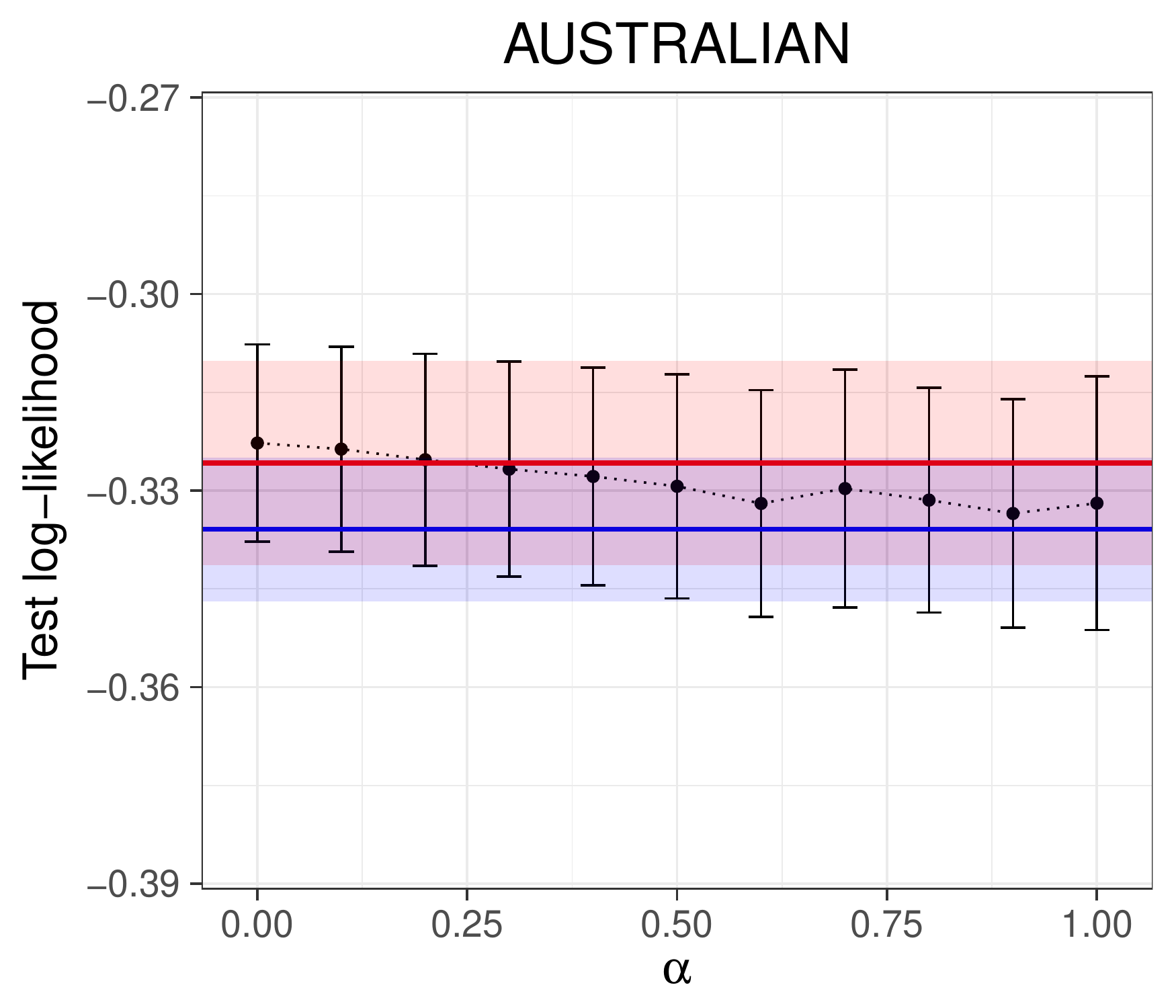} & 
    \includegraphics[width = 0.31\textwidth] {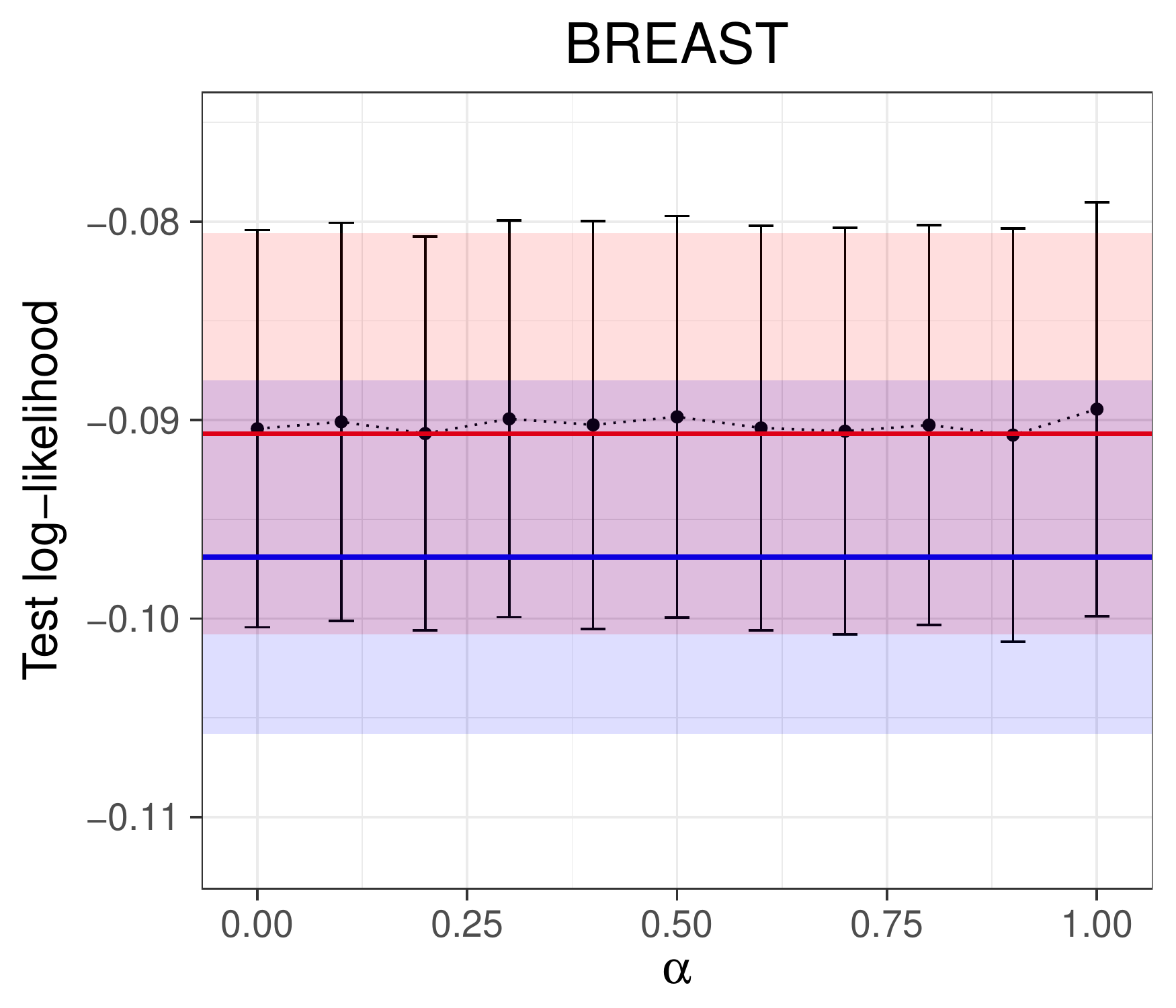} &
    \includegraphics[width = 0.31\textwidth] {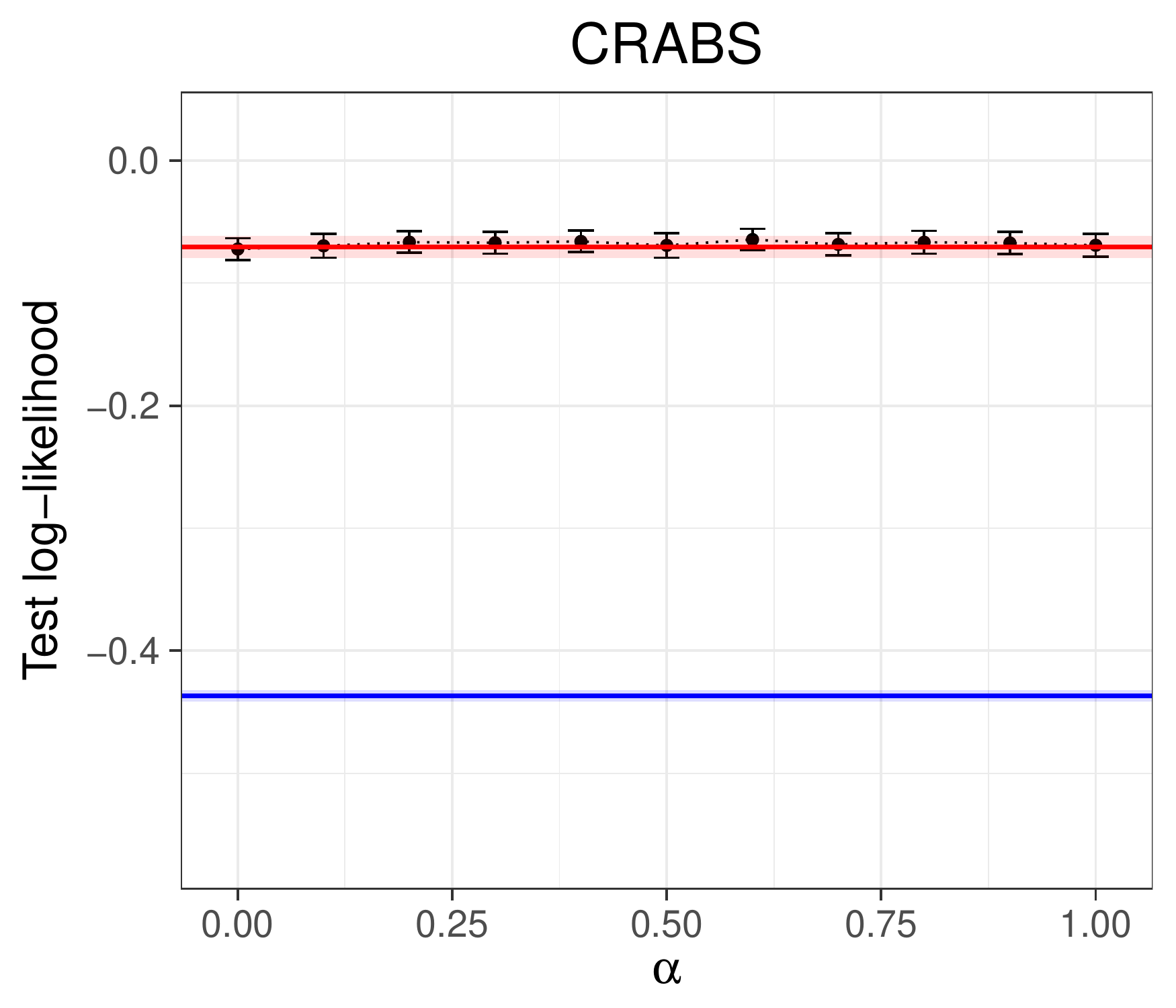} \\ 
    \includegraphics[width = 0.31\textwidth] {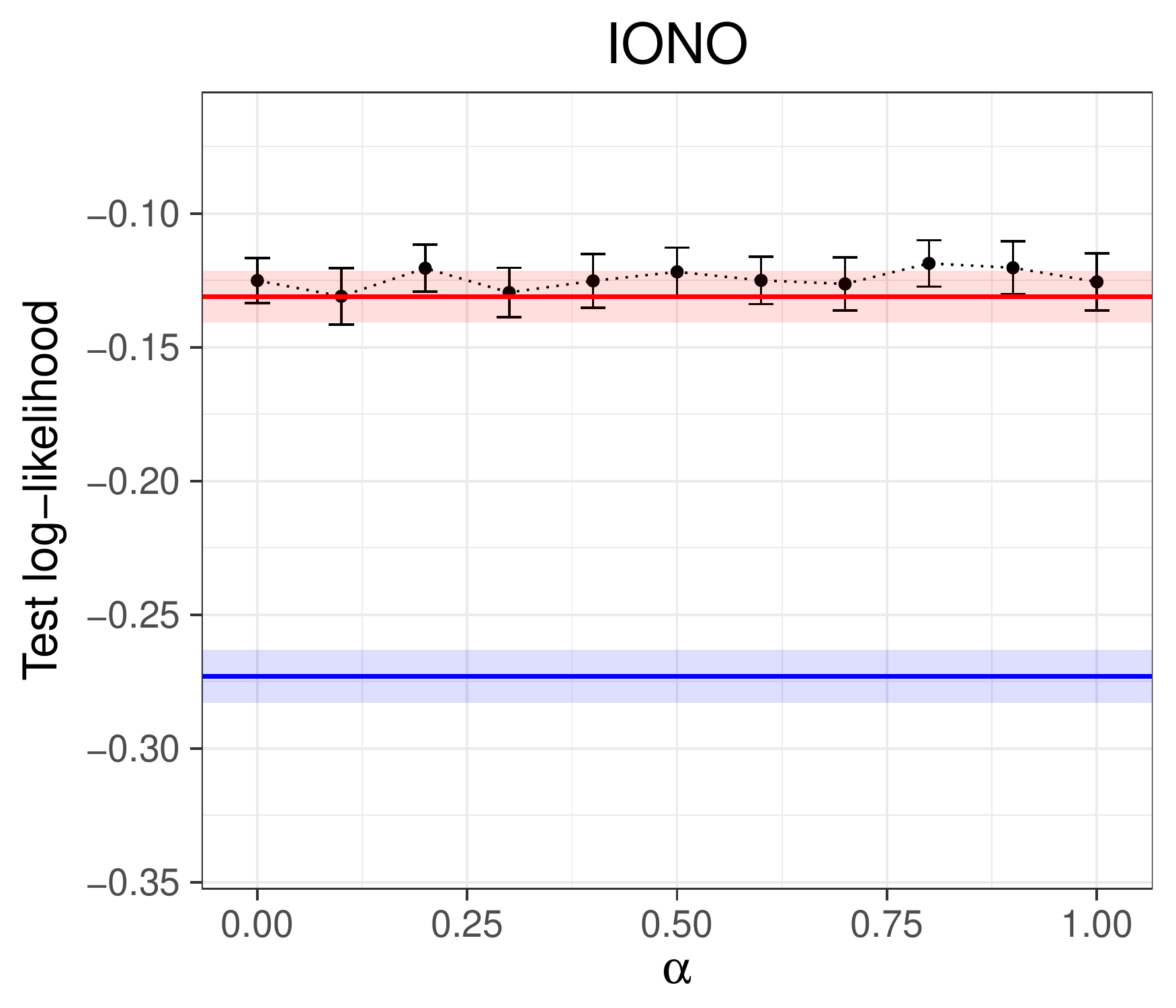} &
    \includegraphics[width = 0.31\textwidth] {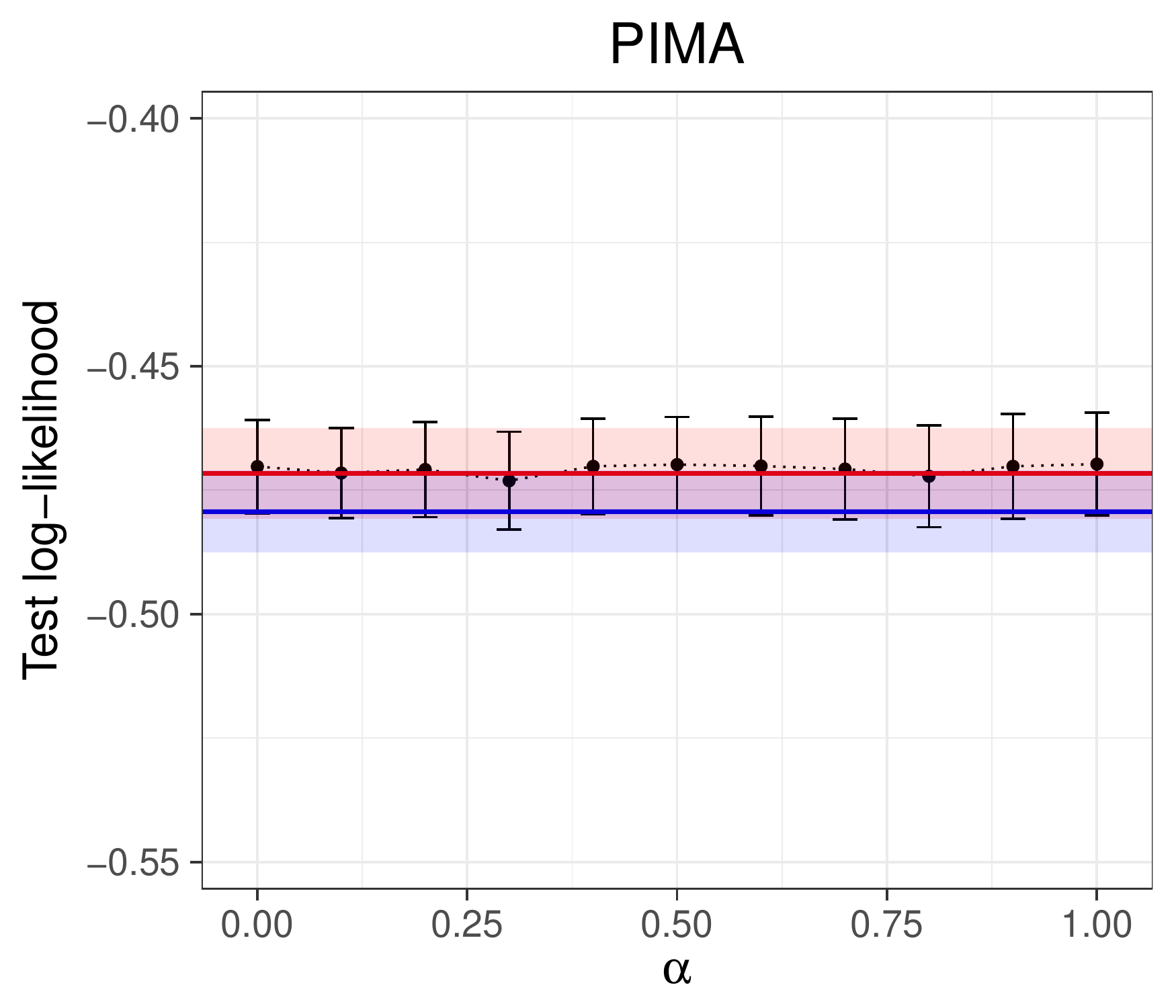} & 
    \includegraphics[width = 0.31\textwidth] {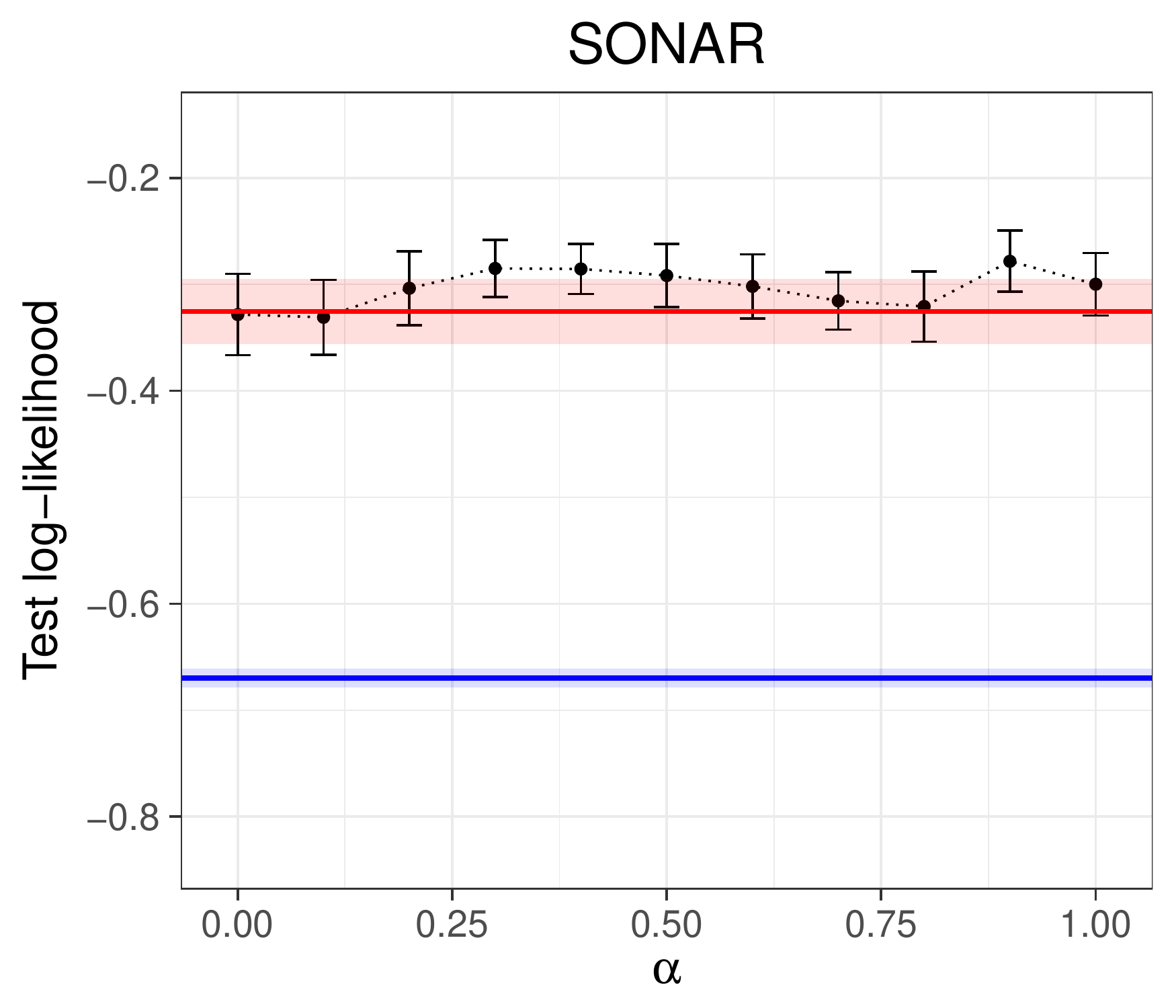} \\
    \multicolumn{3}{c}{\includegraphics[width = 0.34\textwidth] {figures/legend.pdf}}
	\end{tabular}
  \caption{Average results in terms of the test log-likelihood for the different UCI datasets for binary classification to compare the methods. Black represents the performance for our method, AADM, for different values of $\alpha$. Red is the performance of AVB. VI is presented in blue. Higher values are better. Best seen in color.}
  \label{fig:large_figure_experiments_LL_class}
\end{center}
\end{figure}

The average performance of each method on each dataset, in terms of the test log-likelihood, is displayed Figure \ref{fig:large_figure_experiments_LL_class} (and as in the main article, the higher the result the better). When compared to the results of regression it is straightforward to notice that in this case, changing the value of $ \alpha$ does not have a clear impact on the final outcome of the experiment. For these binary classification datasets, the different approximating distributions that are obtained from going over the $[0, 1]$ interval with $\alpha$ seems to not have a distinctive impact on the log-likelihood across all the splits of the datasets, which we will analyze further below. If we take into account previous work, such as the one presented in \cite{bui2017unifying}, we observe that the results achieved by AADM, in terms of the log-likelihood, are either similar or better than the ones presented before. In all experiments, VI comparatively has the worse results, even in cases such as the \emph{Australian} or \emph{Breast} dataset. The width of the error-bars relative to the mean result in the latter case appear to be substantially wider than for the rest of the datasets, although it is just a matter of the scale of the y-axis on the plot. In general it is not possible to extract definite conclusions on whether AVB or AADM perform better overall, although the results in \emph{Sonar} and \emph{Iono} may lead us to think so. Nevertheless, the performance of AADM for all the $\alpha$ values is similar to that of AVB, being almost the same when $\alpha \rightarrow 0$, as expected.

\begin{figure}[]
\begin{center}
	\begin{tabular}{ccc}
    \includegraphics[width = 0.31\textwidth] {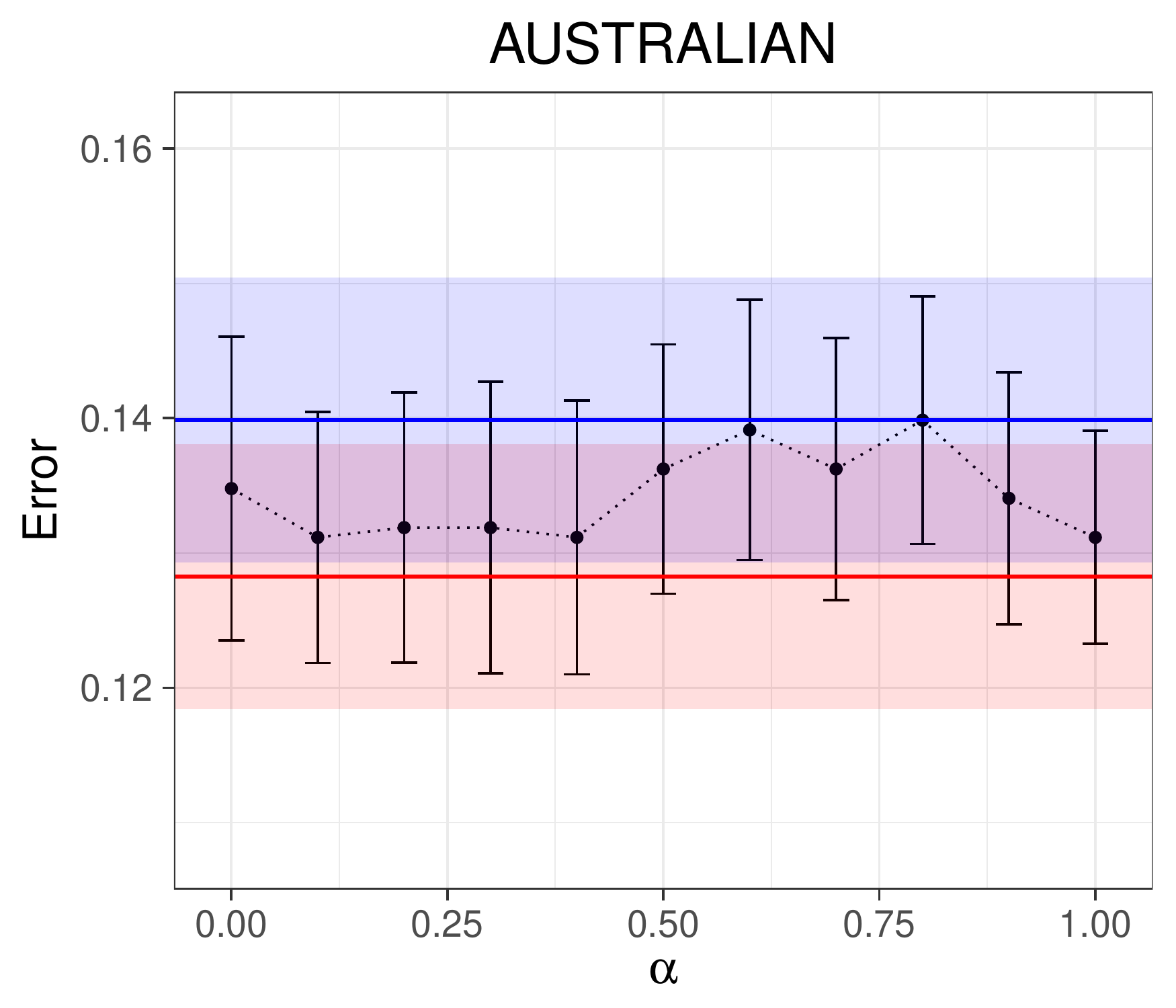} & 
    \includegraphics[width = 0.31\textwidth] {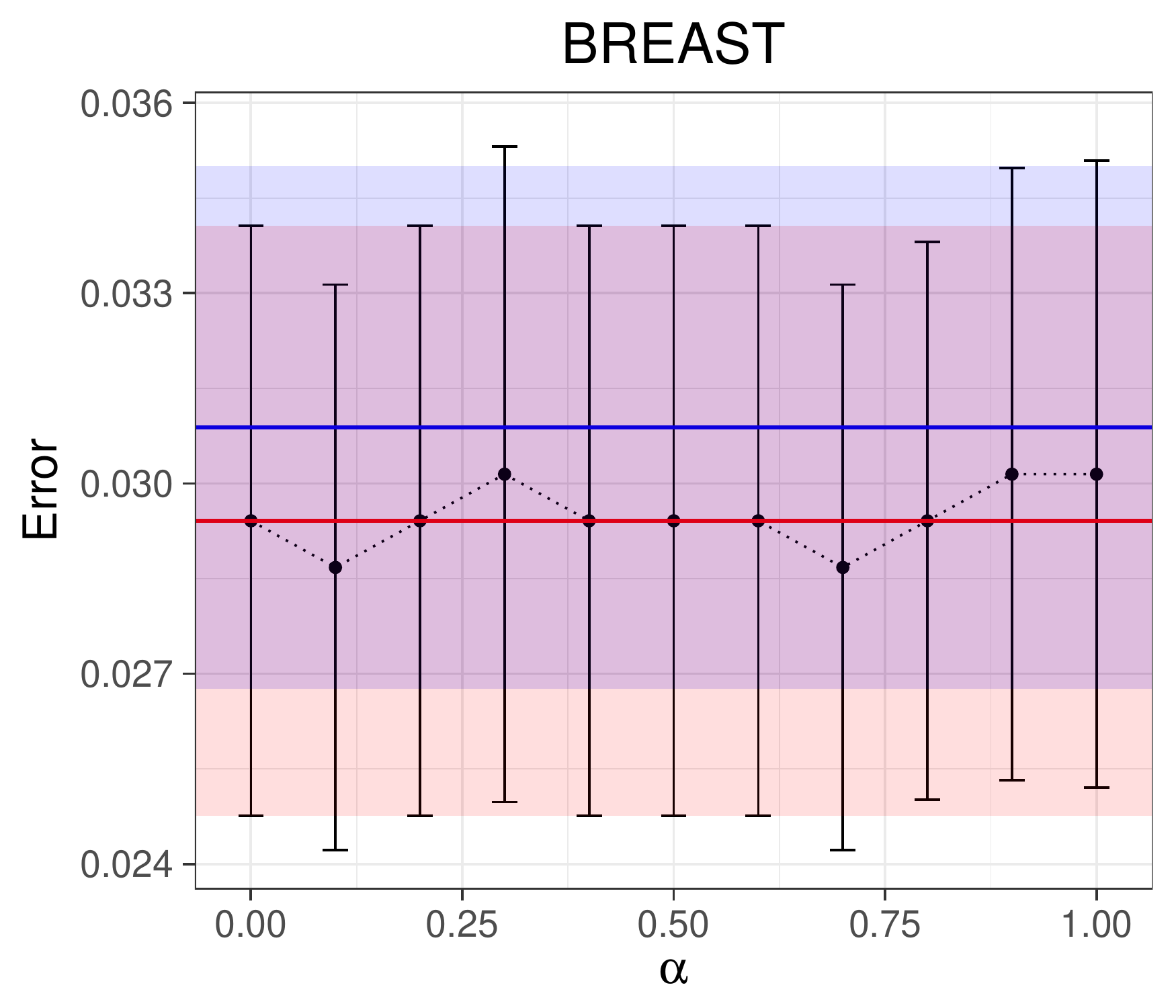} &
    \includegraphics[width = 0.31\textwidth] {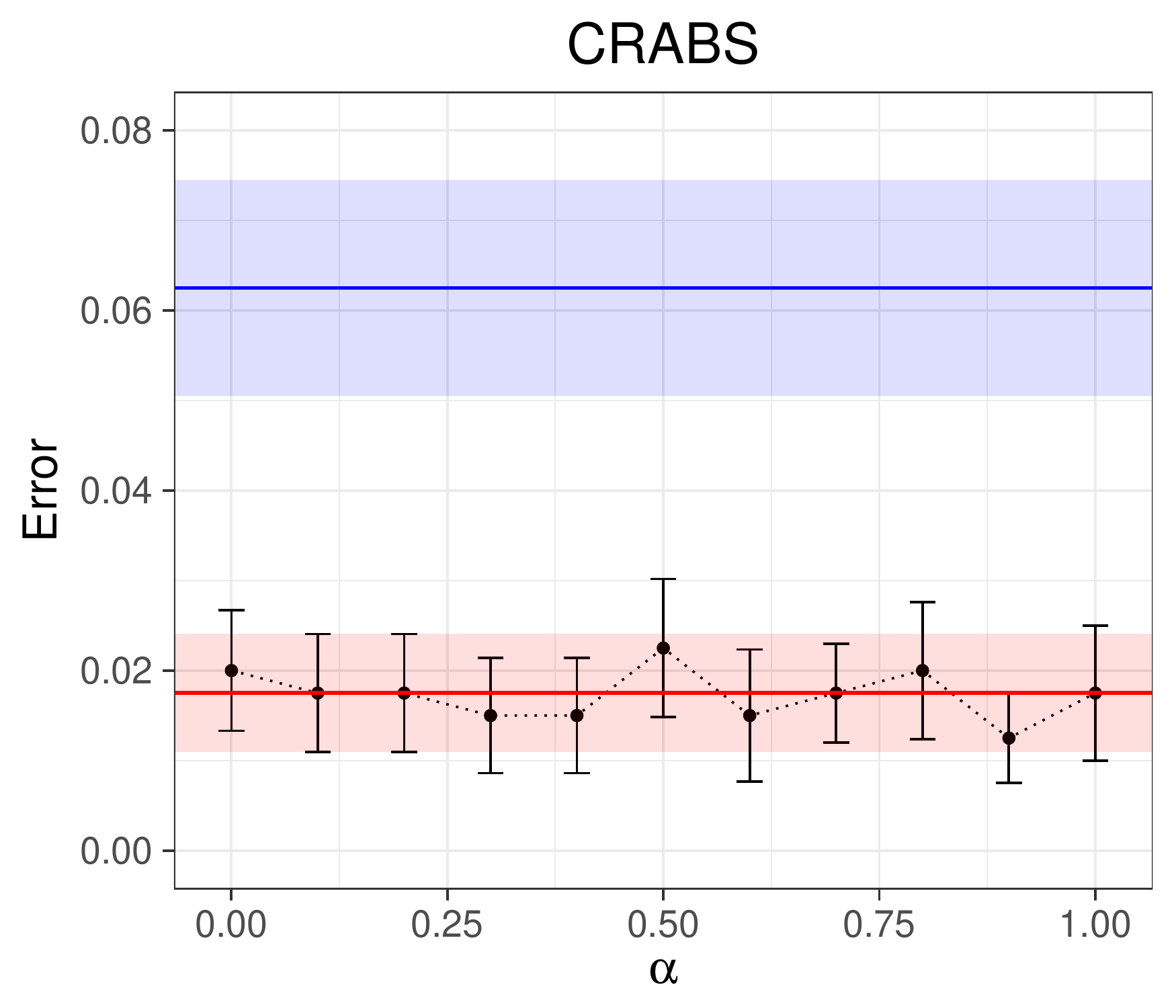} \\
    \includegraphics[width = 0.31\textwidth] {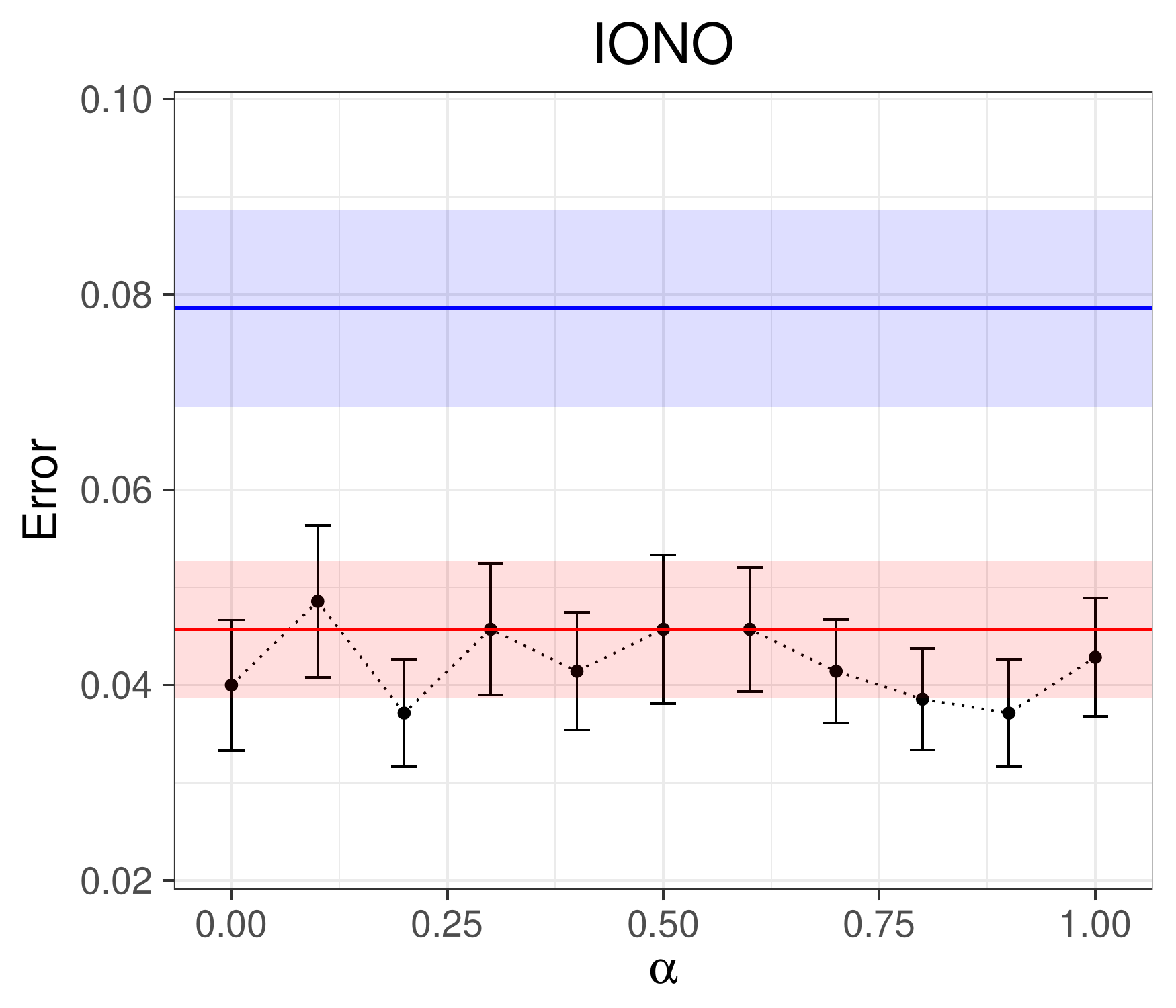} &
    \includegraphics[width = 0.31\textwidth] {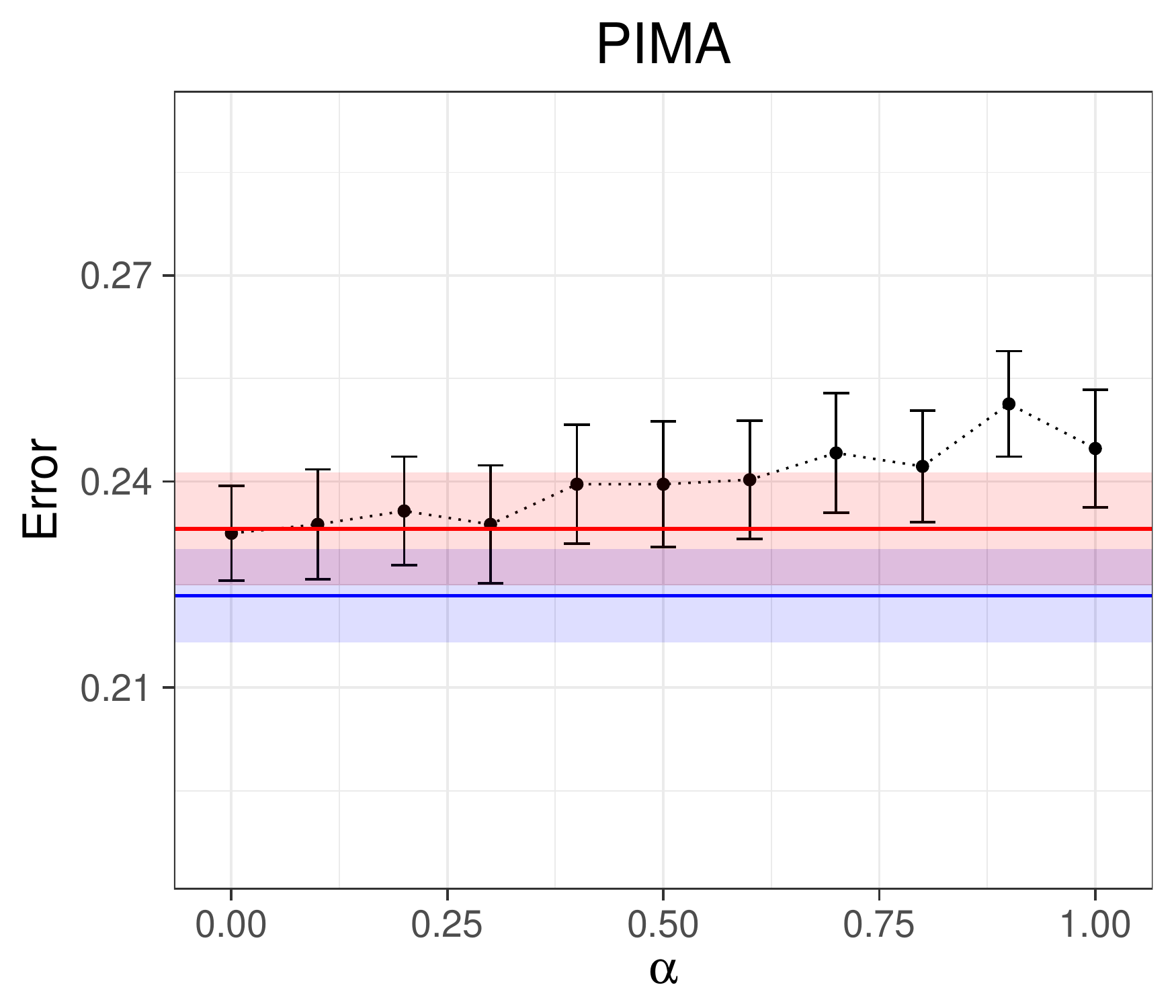} & 
    \includegraphics[width = 0.31\textwidth] {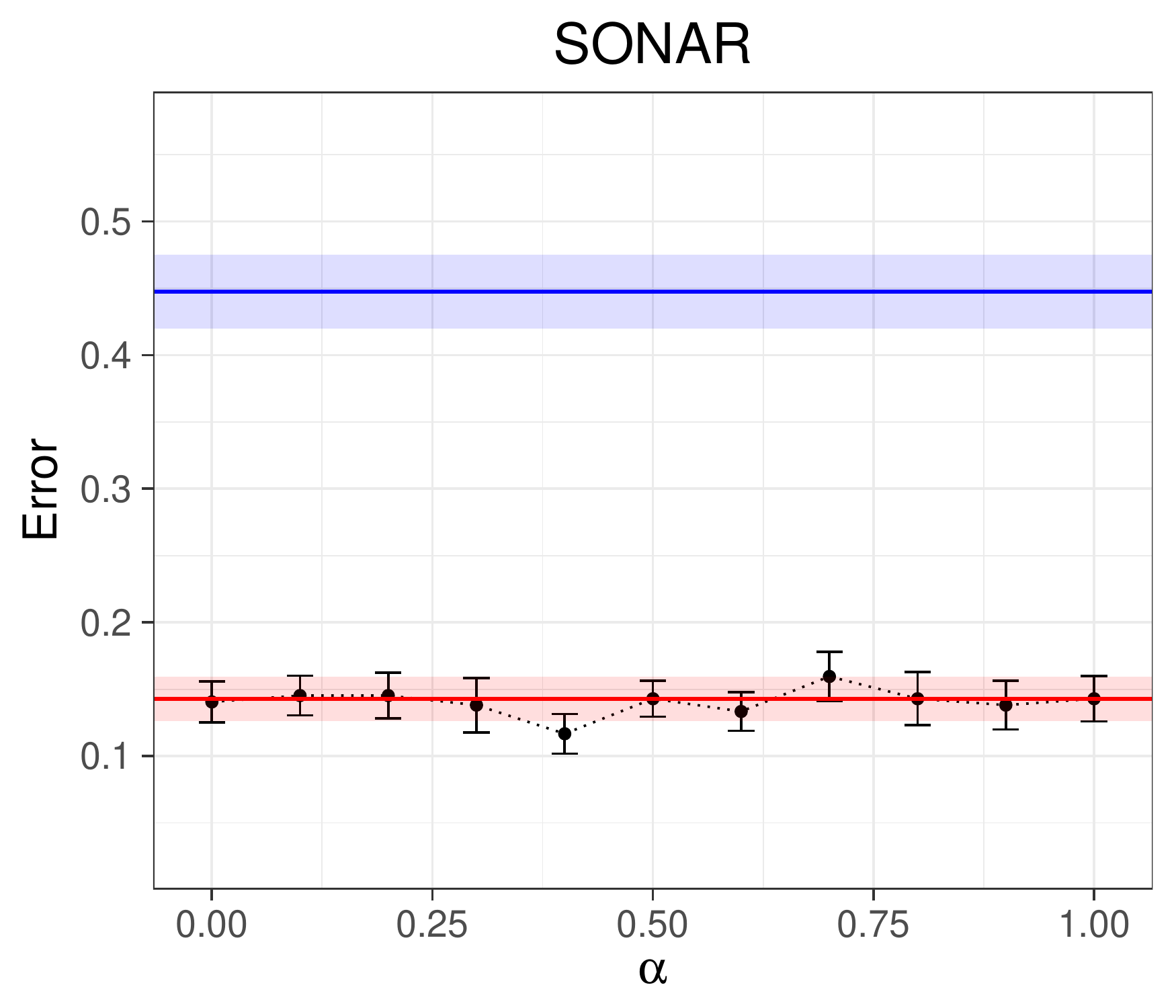} \\
    \multicolumn{3}{c}{\includegraphics[width = 0.34\textwidth]{figures/legend.pdf} }
	\end{tabular}
  \caption{Average results in terms of the classification error rate for the different UCI datasets for comparison of the methods. Black represents the performance for our method, AADM, for different values of $\alpha$. Red is the performance of AVB. VI is presented in blue. Lower values are better. Best seen in color.}
  \label{fig:large_figure_experiments_Error_class}
\end{center}
\end{figure}

In the other hand, the results for the classification error rate are displayed in Figure \ref{fig:large_figure_experiments_Error_class}. With this metric we see the same general trends we observed for the log-likelihood (as in the article, the lower the result the better). Overall, AADM and AVB perform better than VI with the exception of \emph{Pima}, on which VI is able to yield better results although not by a large difference. Overall, VI performs worse across datasets, but yet again it is difficult to conclude whether AVB or AADM perform better than the other since their behaviors are very similar (again, in cases such as \emph{Australian} or \emph{Breast}, the width of the error-bars relative to the mean error is similar to the other experiments, although the scale of the y-axis make them appear a lot wider). Additionally, there is no obvious changing of the behavior of the error with the $\alpha$ value here either. Finally, as expected, AADM and AVB have very similar results when $\alpha \rightarrow 0$ in AADM.

\subsection{Average Rank Results on binary classification tasks}

\begin{center}
\begin{figure}[]
    \includegraphics[width = 0.99\textwidth]{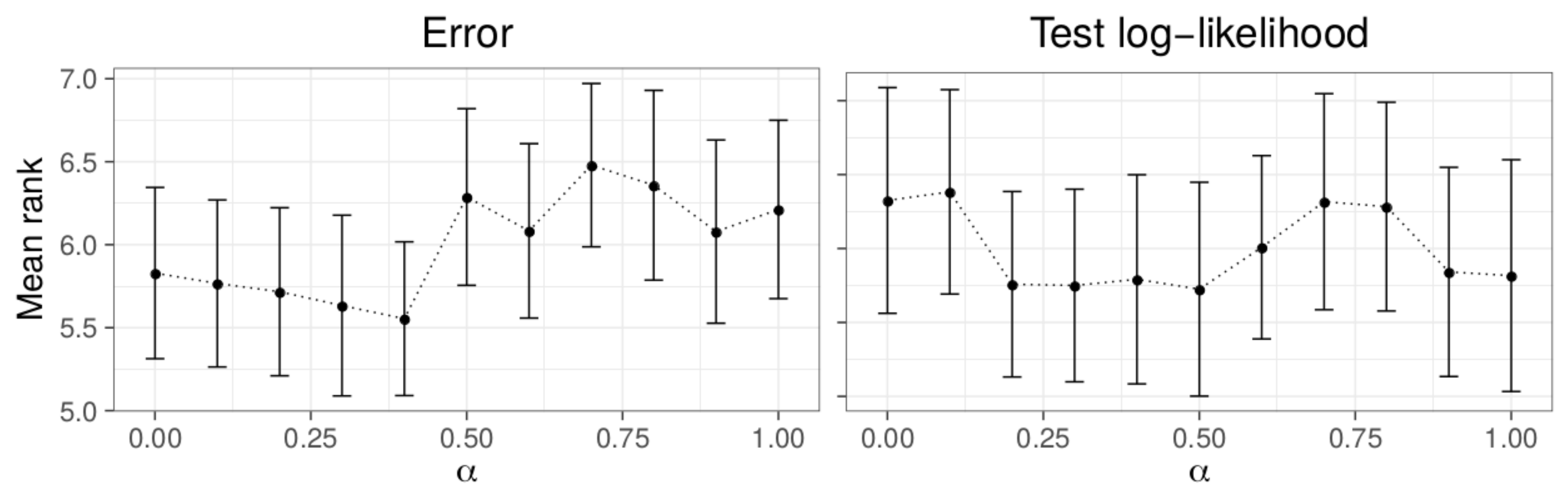}
    \caption{Average rank (the lower the better) for AADM and each value of $\alpha$ in terms of the classification error rate (left) and the test log-likelihood (right) across all the binary classification datasets and splits.}
    \label{fig:final_rank_analysis_class}
\end{figure}
\end{center}

To see if there is a distinct overall change of behavior depending on the choice of $  \alpha$ for AADM we perform the same analysis that was done before for regression: we rank the performance of AADM for each $\alpha$ from best to worse (\emph{e.g.} rank $1$ represents the best result) and compare the ranks for all $\alpha$ to one another, computing the average rank across all binary classification datasets for each final metric. The results both for the error classification rate and the log-likelihood are displayed in Figure \ref{fig:final_rank_analysis_class}.

The results represented here are not as clear as they were in the case of regression. In here, a convex behavior such as the ones obtained in those other cases is not observable, leaving all the $\alpha$ values explored almost in the same performance in terms of their rank when compared to each other, as seen in Figure \ref{fig:final_rank_analysis_class}. In the case of the classification error rate we can say that there is a slight improvement for $\alpha$ values lower than $0.5$, as can be seen in the left-side plot. However, when comparing log-likelihoods there is no clear behavior that can be extracted from the experiments. Analyzing the resulting data of the experiments we have come to the conclusion that AADM may need more data to make the best of its generalized approach for modeling the posterior distribution of the parameters on the NN system. We have performed thorough analysis on these results and its possible causes, and we find it is most likely due to the amount of data available, as well as maybe noise in the data itself. The general trend here obtained suggests there is room for improvement and analysis in order to discern if there are any overall better values for $\alpha$ or not, although this would be the subject of further research. However, even though not a simple convex behavior is observed such as the one in regression, the performance achieved across all the $\alpha$ values tested here remains comparable to that of AVB. Therefore, regarding binary classification tasks AADM operates at the level of current state-of-the-art methods. Meanwhile, it is able to improve on them on regression as we describe in the main part of the article.

\section*{References}

\bibliography{bibfile_supplementary}